\documentclass[preprint,12pt]{elsarticle}
\usepackage{lineno,hyperref}
\usepackage{natbib}
\usepackage{geometry}
\usepackage{fleqn}
\usepackage{graphicx}
\usepackage{newtxtext,newtxmath}
\usepackage{hyperref}
\usepackage{booktabs}
\usepackage{bm}
\usepackage{amsmath}
\usepackage{enumitem}
\usepackage{esvect}
\usepackage{soul,xcolor}
\usepackage{caption}
\usepackage{graphicx,subcaption}
\modulolinenumbers[5]

\graphicspath{{./Figures/}}
\journal{Nuclear Engineering and Design}

\begin{document}

\begin{frontmatter}

\title{\large{Deep Generative Modeling-based Data Augmentation with Demonstration using the BFBT Benchmark Void Fraction Datasets}}

\author[NCSU]{Farah Alsafadi}

\author[NCSU]{Xu Wu\corref{mycorrespondingauthor}}
\cortext[mycorrespondingauthor]{Corresponding author}
\ead{xwu27@ncsu.edu}

\address[NCSU]{Department of Nuclear Engineering, North Carolina State University    \\ 
	Burlington Engineering Laboratories, 2500 Stinson Drive, Raleigh, NC 27695 \\}

\begin{abstract}

Deep learning (DL) has achieved remarkable successes in many disciplines such as computer vision and natural language processing due to the availability of ``big data''. However, such success cannot be easily replicated in many nuclear engineering problems because of the limited amount of training data, especially when the data comes from high-cost experiments. To overcome such a \textit{data scarcity} issue, this paper explores the applications of deep generative models (DGMs) that have been widely used for image data generation to scientific data augmentation. DGMs, such as generative adversarial networks (GANs), normalizing flows (NFs), variational autoencoders (VAEs), and conditional VAEs (CVAEs), can be trained to learn the underlying probabilistic distribution of the training dataset. Once trained, they can be used to generate synthetic data that are similar to the training data and significantly expand the dataset size. By employing DGMs to augment TRACE simulated data of the steady-state void fractions based on the NUPEC Boiling Water Reactor Full-size Fine-mesh Bundle Test (BFBT) benchmark, this study demonstrates that VAEs, CVAEs, and GANs have comparable generative performance with similar errors in the synthetic data, with CVAEs achieving the smallest errors.
%While the NFs model yields higher error values, they still fall within an acceptable range.
The findings shows that DGMs have a great potential to augment scientific data in nuclear engineering, which proves effective for expanding the training dataset and enabling other DL models to be trained more accurately.

\end{abstract}


\end{frontmatter}
%\linenumbers

%%%%%%%%%%%%%%%%%%%%%%%%%%%%%%%%%%%%%%%%%%%%%%%%%%%%%%%%%%%%%%%%%%%%%%%%%%%%%%%%%%%%%%%%
%%%%%%%%%%%%%%%%%%%%%%%%%%%%%%%%%%%%%%%%%%%%%%%%%%%%%%%%%%%%%%%%%%%%%%%%%%%%%%%%%%%%%%%%
\section{Introduction}

%%%%%%%%%%%%%%%%%%%%%%%%%%%%%%%%%%%%%%%%
\subsection{The data scarcity issue in nuclear engineering}

Deep learning (DL) has achieved remarkable successes in many disciplines such as computer vision and natural language processing due to the availability of ``big data'', which encompasses a large number of images, audios, videos, spoken words, etc. However, when it comes to scientific data in nuclear engineering, DL algorithms that rely on deep neural networks (DNNs) face challenges that necessitate task-specific modifications. This is primarily due to the fact that many nuclear engineering problems usually has ``small data'', especially for cases when the data is obtained from costly experiments. While large amounts of data may be available for certain components like pumps, turbines, and pipes due to sensor networks, there is a significant scarcity of data for critical aspects such as void fraction, departure from nucleate boiling, and thermo-physical properties of molten salts. Particularly concerning is the lack of data for advanced reactor design and safety analysis, raising challenges for utilizing machine learning (ML) in licensing analyses of advanced nuclear reactors. Therefore, it is crucial to develop reliable techniques for data augmentation to expand the training dataset. This approach is essential for overcoming over-fitting issues and enabling DL algorithms to achieve better performance.

%%%%%%%%%%%%%%%%%%%%%%%%%%%%%%%%%%%%%%%%
\subsection{Data augmentation with deep generative models}

Data augmentation \cite{shorten2019survey} is a fundamental component of data science that involves using computational techniques to increase the amount of data. It can be achieved by creating modified copies of existing data or generating synthetic data that resembles the behavior of the original data. For example, in handwriting recognition algorithms using the MNIST dataset, a training image of a digit can be rotated by a small angle to create a slightly modified version of the same digit. Image data augmentation commonly involves techniques like flipping, rotation, zooming, cropping, and color adjustment, which can enhance performance in various learning tasks beyond just handwriting recognition. The key idea is to expand the training data by applying operations that capture real-world variations. However, these traditional data augmentation techniques cannot be directly applicable to non-image scientific data, including those in nuclear engineering. To address the issue of data scarcity, Deep generative models (DGMs) offer a promising solution. \textit{DGMs are usually DNNs trained to model the underlying probabilistic distribution of a dataset, which allows them to generate synthetic data that follows the same distribution as the training data}. By leveraging well-trained DGMs, it becomes possible to significantly increase the size of an existing dataset by simply generating samples from the learned distribution.

Since data augmentation with DGMs is still a relatively new research area in nuclear engineering, this study aims to explore the effectiveness of several DGMs for expanding a dataset. Specifically, we investigate models including generative adversarial networks (GANs), real-valued non-volume preserving (real NVP) normalizing flows (NFs), variational autoencoders (VAEs), and a variant of VAE called conditional VAE (CVAEs). CVAE operates similarly to VAEs but with the additional capability of generating specific data based on provided labels. These models and their variations have achieved extraordinary successes in image data augmentation in computer vision and natural language processing. By evaluating the performance of these models using scientific data in a nuclear engineering problem, we can determine their suitability for generating synthetic data to expand the existing datasets. Each DGM technique employs a distinct approach to learn the distribution of the training data and generate synthetic data. In the following we will provide a brief overview of each DGM method and a short review of its applications.

%%%%%%%%%%%%%%%%%%%%%%%%%%%%%%%%%%%%%%%%
\subsection{Overview and applications of GANs}

GANs have received much attention after the seminal work by Goodfellow et al. in 2014 \cite{goodfellow2014generative}. This model consists of two neural network models, namely the \textit{generator} and the \textit{discriminator}. The generator's task is to learn the data distribution and generate data that resembles the training dataset, while the discriminator acts as a classifier, trained to correctly label both real and synthetic data. In the training of GANs the two models compete in a minimax two-player game, where the generator generates samples to deceive the discriminator, while the discriminator strives to accurately classify the data. The training process continues until the generator generates plausible samples that the discriminator can no longer distinguish from real data.

To enhance their capabilities for specific tasks, several task-specific GANs variants have been introduced. Conditional GANs (CGANs) \cite{mirza2014conditional} were developed to generate targeted data by conditioning the training process on specific labels. Deep tensor GANs (TGANs) were proposed in \cite{ding2019tgan} to generate high quality images by leveraging tensor structures. GANs have been used for generating facial images \cite{yin2017semi} as well as videos \cite{vondrick2016generating}. GANs have also been utilized for sequential data applications. For instance, researchers proposed RankGAN \cite{lin2017adversarial} to generate high-quality language descriptions, and it was employed to generate neural dialogue as well \cite{li2017adversarial}. In \cite{hu2017generating}, GANs were applied in cyber security field to identify malware. GANs have also been successful in many other fields. For example, in the medical field, they were used in generating DNA sequences and tuning them to have specific properties \cite{killoran2017generating}. Despite the advancements made in various fields, the application of GANs in nuclear engineering is still limited. A notable development in this area is an algorithm called BubGAN \cite{fu2019bubgan}, which was designed to generate realistic images of bubbly flow by conditioning the generation process on bubble features. The study demonstrated that this generative algorithm effectively reduces the cost of labeling for bubble detection and segmentation algorithm development. Note that this work still used image data for training.

%%%%%%%%%%%%%%%%%%%%%%%%%%%%%%%%%%%%%%%%
\subsection{Overview and applications of NFs}

NFs \cite{rezende2015variational} differ from GANs in their approach to learning the training data distribution. NFs utilize invertible functions, also known as bijective functions, and the change-of-variables rule to transform the complex distribution of the data into a simpler distribution. This is known as the \textit{training direction}. By repeatedly applying the invertible functions and the change-of-variables rule in the training phase, NFs learn the data distribution by transforming the complex input data into the assumed simpler distribution, typically a normal distribution. Neural networks are used to implement the invertible mappings. To generate new samples, the process is reversed, known as the \textit{generation direction}. The network takes samples from the simpler distribution and applies an inverse transformation to generate data samples that resemble the original complex distribution of the training data. This allows NFs to generate realistic samples from the learned distribution. Unlike GANs and VAEs, NFs can approximate the maximum likelihood exactly. Hence, they offer a more stable training process and efficient sampling of new data. NF-based techniques have been successfully applied in various fields, including but not limited to image generation \cite{ho2019flow++}, audio and video generation \cite{ping2020waveflow} \cite{kumar2019videoflow}, as well as graph generation \cite{madhawa2019graphnvp}. %Moreover, NFs have found applications in physics. For example, equivariant flow-based sampling has been used for lattice gauge theory \cite{kanwar2020equivariant}, enabling efficient sampling of configurations. Additionally, NFs have been employed to sample configurations for multi-body systems with symmetric energies \cite{kohler2019equivariant}.

%%%%%%%%%%%%%%%%%%%%%%%%%%%%%%%%%%%%%%%%
\subsection{Overview and applications of VAEs}

Unlike GANs and NFs, VAEs employ a unique approach that combines variational inference and DL to approximate the underlying probability distribution of the training data. VAEs consist of three main components: the encoder, latent space, and the decoder. The key idea behind VAEs, which makes them suitable for data generation unlike regular autoencoders, is the encoding process. Instead of encoding the input into a fixed vector, the encoder encodes the input data into a distribution. Thus, instead of having a single vector for the latent space, it will have two vectors, one for the mean and one for the standard deviation of the encoded distribution. The decoder takes samples from the latent space and reconstructs the original input by decompressing the data.
%During the training process, the encoder encodes the input data, and the decoder aims to reconstruct the original input.
The reconstruction loss is calculated to provide feedback for model improvement, ensuring a better reconstruction of the original data and preserving important features that may have been lost during the encoding process. After the model is properly trained, the decoder alone is utilized to generate synthetic data by taking samples from the latent space.

The standard VAE model has a limitation in generating specific data at specified locations within the training data domain. To address this limitation, a variant of VAEs called conditional VAEs (CVAEs) was proposed \cite{sohn2015learning}. CVAEs operate similarly to VAEs but utilize additional data for conditioning. During the training phase, instead of solely relying on the training data, CVAEs incorporate labels alongside the data to train both the encoder and the decoder. This incorporation of labels enables the decoder to learn how to generate data that corresponds to specific labels, allowing for more targeted data generation.

VAEs have been utilized for collaborative filtering in implicit feedback scenarios \cite{liang2018variational}. Additionally, VAEs have shown effectiveness in improving product quality prediction by generating artificial quality values for training \cite{lee2023developing}, as well as in computer vision tasks like generating static images \cite{walker2016uncertain} and enhancing image super resolution \cite{sonderby2016amortised}. These examples highlight the effectiveness of VAEs in various fields.

%%%%%%%%%%%%%%%%%%%%%%%%%%%%%%%%%%%%%%%%
\subsection{Overview of this work}

In this paper, we applied the above DGMs for augmenting the steady-state void fraction data obtained from TRACE simulations based on the NUPEC Boiling Water Reactor Full-size Fine-mesh Bundle Test (BFBT) benchmark \cite{neykov2005nupec}. The DGMs used in this study include GANs, real NVP NFs, VAEs, and CVAE. Our objective was to generate synthetic samples using these models and evaluate their credibility. The major contribution of this work is the novel application of DGMs that have been successful for image data generation to scientific data in a nuclear engineering problem. Note that the primary reason for us to demonstrate the DGMs using computer model simulation data, instead of real experimental data, is that we can run the computer model at the inputs of the synthetic data in order to evaluate the error in the generated data. This would be impossible if real experimental dataset is used, as we cannot perform new experiments at the potentially random inputs of the generated data. The results indicate that all four models were able to produce plausible samples, with VAEs, CVAEs, and GANs exhibiting comparable smaller error values than NFs. Among them, CVAEs achieved the smallest errors.
%Although NFs had higher error values, they still fell within an acceptable range.
The findings demonstrate that DGMs have the potential to generate synthetic scientific data that can effectively expand the training dataset, thus improving the accuracy of ML/DL models trained on the expanded dataset.

The organization of the remaining sections in this paper is as follows: Section \ref{sec:Problem_def} presents the problem definition, provides an overview of the training data, and discusses the validation process. In Section \ref{sec:Methodologies}, we provide a self-contained description of the methodologies and characteristics of the four DGMs employed in this study. Section \ref{sec:Results} presents the results of data augmentation using the four models, along with a validation of the synthetic data through a comparison with TRACE simulation results. Finally, Section \ref{sec:Conclusions} concludes the paper and highlights potential avenues for future research.

%%%%%%%%%%%%%%%%%%%%%%%%%%%%%%%%%%%%%%%%%%%%%%%%%%%%%%%%%%%%%%%%%%%%%%%%%%%%%%%%%%%%%%%%
%%%%%%%%%%%%%%%%%%%%%%%%%%%%%%%%%%%%%%%%%%%%%%%%%%%%%%%%%%%%%%%%%%%%%%%%%%%%%%%%%%%%%%%%
\section{Problem Definition}
\label{sec:Problem_def}

%%%%%%%%%%%%%%%%%%%%%%%%%%%%%%%%%%%%%%%%
\subsection{Test problem and training data generation}

The training dataset is derived from TRACE \cite{USNRC2014TRACE} simulations of steady-state void fractions based on the BFBT benchmark \cite{neykov2005nupec}.  The main reason for using TRACE simulated data instead of the experimental data in this work is that, it is possible to assess the accuracy of the generated data by running TRACE at the inputs of the generated samples, and comparing the outputs of the generated samples with simulated outputs. If the experimental data is used, this validation process becomes unfeasible due to the fact that we cannot perform new experiments at the potentially random inputs of the generated data. Therefore, in this work we will only focus on demonstration using purely computational data, and reserve tests of DGMs using the experimental data for future work.

The BFBT facility replicates a full-scale Boiling Water Reactor fuel assembly and enables the measurement of void fraction distribution at four axial locations. To obtain these measurements, X-ray CT scanner and X-ray densitometer were employed. The measured void fractions at the four axial positions will hereafter be referred to as \texttt{VoidF1}, \texttt{VoidF2}, \texttt{VoidF3}, and \texttt{VoidF4} from bottom to top, respectively. The BFBT benchmark experiments investigated five different assembly configurations, each characterized by distinct geometries and power profiles. Within each assembly configuration, multiple tests were conducted by varying four design variables: pressure, mass flow rate, power, and inlet temperature. For the purpose of illustration in this work, assembly 4101 test number 55 was selected due to its utilization as a baseline TRACE model in the BFBT benchmark.

The training dataset is generated using the system thermal-hydraulics code TRACE (version 5.0 Patch 4) \cite{USNRC2014TRACE}. TRACE offers the capability to perturb 36 physical model parameters (PMPs) from the input deck. In a previous study \cite{wu2018inverse} that involved sensitivity analysis and inverse uncertainty quantification of TRACE based on the BFBT benchmark, the most sensitive PMPs were identified, as listed in Table \ref{table:TRACE-model-parameters}. To create the training dataset, the TRACE code is executed 200 times, with the five PMPs randomly sampled within a uniform range of $(0.0,5.0)$. Each TRACE run takes about 40 seconds, and it generates void fraction values for the four axial locations. This dataset, with the five PMPs as inputs and the four void fractions as outputs, will be used to train the DGMs. Note that all the selected DGMs are composed of neural networks that need a decent amount of data to train. In this work a small dataset size of only 200 samples is intentionally chosen in order to test the capability of the DGMs to learn the data distributions under data scarcity.

\begin{table}[htbp]
	\footnotesize
	\captionsetup{justification=centering}
	\caption{List of TRACE PMPs (multiplicative factors) that are significant to the BFBT benchmark \cite{wu2018inverse}.}
	\label{table:TRACE-model-parameters}
	\centering
	\begin{tabular}{l c c c c}
	   \toprule
          Physical model parameters in TRACE & Symbols & Uniform ranges & Nominal values\\
          \midrule
          Single phase liquid to wall heat transfer coefficient& \texttt{P1008} & (0.0, 5.0) & 1.0\\
          Subcooled boiling heat transfer coefficient& \texttt{P1012}   & (0.0, 5.0) & 1.0\\
          Wall drag coefficient & \texttt{P1022} & (0.0, 5.0) & 1.0\\
          Interfacial drag (bubbly/slug Rod Bundle) coefficient & \texttt{P1028} & (0.0, 5.0) & 1.0\\
          Interfacial drag (bubbly/slug Vessel) coefficient & \texttt{P1029} & (0.0, 5.0) & 1.0\\
        \bottomrule
	\end{tabular}
\end{table}

%%%%%%%%%%%%%%%%%%%%%%%%%%%%%%%%%%%%%%%%
\subsection{DGMs training and validation}

Once the DGMs are trained, new samples can be generated as the synthetic data. Each sample is a \textit{nine-dimensional vector, including five PMPs (TRACE inputs) and four void fractions (TRACE outputs)}. Note that unlike conventional regression-based ML/DL models that approximate the input-output relationship of the training data, DGMs are generative models that learn the data distributions. They do not explicitly learn the input-output relationship, nevertheless, in each sample vector the output elements correspond to the input elements. The data generation process is extremely fast because it only involves sampling from the learned distribution and evaluate through the DGMs, which are mostly neural networks. This proves advantageous in situations where simulations are computationally expensive or when data is derived from costly experiments. Moreover, this approach can be viewed as analogous to ``surrogate modeling'', which employs a DNN to learn the input-output relationship from a training dataset and generate new samples. The key distinction between a DNN-based DGM and a DNN-based surrogate model lies in the former's ability to learn the underlying probabilistic distributions of the data, whereas the latter constructs a black-box surrogate without considering the distributions. Hence, DGMs are generally recognized as a form of probabilistic ML.

To perform validation, 500 samples are generated by each DGM. These generated samples firstly undergo pre-processing to ensure consistency with the training data. Specifically, any PMP values outside the range of $(0.0,5.0)^5$ are eliminated, because the corresponding samples are outside of the validity of the training dataset. Subsequently, the validation data is obtained by running TRACE at the five-dimensional inputs of the generated samples. The resulting four-dimensional outputs will be compared with the void fraction values in the generated samples. The errors between the void fractions will be used to evaluate how well the generated values align with the actual data obtained from the TRACE code. One might have questions concerning eliminating the samples outside of the training domain $(0.0,5.0)^5$ - (1) aren't DGMs expected to generate data out of the training domain? (2) what is the point of only keeping samples inside the training data? We argue that, in this preliminary work we do not expect to achieve powerful methods in a single step. The goal of this work is to demonstrate the DGMs and learn what adjustments need to be made in order for them to work on experimental data. Expanding the dataset size without expanding the data domain is still useful, as it is generally true that ML/DL models are more accurate when trained with more data (of course this also depends on the problem itself and the ML/DL model architecture). Developing DGMs that can be used to expand the dataset beyond the training domain is reserved for future work.

%%%%%%%%%%%%%%%%%%%%%%%%%%%%%%%%%%%%%%%%%%%%%%%%%%%%%%%%%%%%%%%%%%%%%%%%%%%%%%%%%%%%%%%%
%%%%%%%%%%%%%%%%%%%%%%%%%%%%%%%%%%%%%%%%%%%%%%%%%%%%%%%%%%%%%%%%%%%%%%%%%%%%%%%%%%%%%%%%
\section{Methodologies}
\label{sec:Methodologies}

To make this paper self-contained, in this section we present a brief introduction of the theories of GANs, NFs, VAEs and CVAEs. Since these models have mainly been used for dealing with data (images, videos, etc.) in computer vision and natural language processing, the main novelty of this work is to develop codes to apply them for scientific data and demonstrate them in a nuclear thermal-hydraulics problem with a widely used benchmark.

%%%%%%%%%%%%%%%%%%%%%%%%%%%%%%%%%%%%%%%%
\subsection{A brief theory of GANs}

GANs consist of two neural network models, namely the generator $G$ and the discriminator $D$. The generator's role is to capture the data distribution and generate synthetic data that resembles the training data. Conversely, the discriminator acts as a classifier, trained to differentiate between the synthetic data generated by $G$ and the real training data. These two models are trained adversarially in a minimax two-player game, where $G$ aims to maximize the probability of $D$ making mistakes, and $D$ tries to not get fooled by $G$ and to classify the synthetic/real data correctly. The competition between these models leads to improvement for both. The training process stops when $G$ successfully fools $D$ by generating samples that closely resemble the real data. At this point $D$ can no longer distinguish between real and synthetic samples.

\begin{figure}[!ht]
    \centering
	\captionsetup{justification=centering}
    \includegraphics[width=0.85\textwidth]{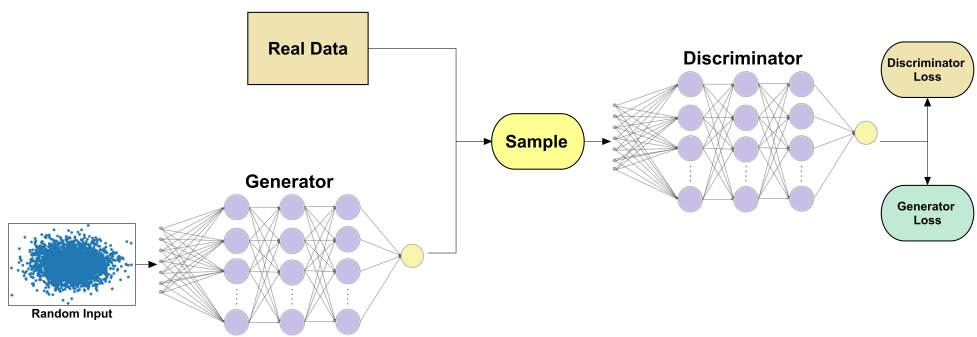}
    \caption{A typical structure of GANs.}
    \label{fig:GAN-diagram}
\end{figure}

Figure \ref{fig:GAN-diagram} illustrates the typical architecture of GANs. The generator receives random inputs $\mathbf{z}$ drawn from a prior distribution and transforms them into samples $G(\mathbf{z})$. These synthetic samples, along with real data samples, are combined and presented to $D$. $D$'s role is to classify these samples as either synthetic or real data. Both models undergo a backpropagation process for learning. During the training of $D$, it is penalized through its loss for classifying the samples incorrectly \cite{goodfellow2020generative}. The feedback from $D$'s loss is then used to update $D$ while keeping $G$ unchanged.

The generator $G$ is trained using feedback from its loss connected to the discriminator $D$, employing backpropagation while keeping $D$ unchanged. This iterative process continues until the discriminator can no longer improve itself. Throughout the training, $G$ learns the distribution of the real data and generates similar data. As $G$ captures the distribution, $D$'s accuracy drops to $50\%$ for classifying the data, which is equivalent to random guessing. This state is referred to as equilibrium, marking the termination of the training process. Once the training process is complete, $G$ can be used alone to generate new data by taking an arbitrary random vector as input and generate a sample vector that consists of both the inputs and outputs in the training dataset.

\begin{table}[htbp] 
	\footnotesize
	\captionsetup{justification=centering}
	\caption{Definitions of mathematical symbols for GANs.}
	\label{table:GANs_symbols_definitions}
	\centering
	\begin{tabular}{cl}
		\toprule
	    Symbols & Meanings \\ 
		\midrule
		$\mathbf{z}$  & Random input vector to the generator  \\
		$\mathbf{x}$  & Samples from real dataset    \\
		$p_Z (\mathbf{z})$ &  Distribution of $\mathbf{z}$\\
	    $p_X (\mathbf{x})$ &  Distribution of $\mathbf{x}$\\
        $p_G$ & Generator's data distribution over data $\mathbf{x}$\\
	    $G(\mathbf{z})$ & Synthetic samples generated by $G$ \\
		$D(\mathbf{x})$ & Discriminator output probability over real samples \\
		$D(G(\mathbf{z}))$ & Discriminator output probability over synthetic samples \\
        $\mathcal{L} (G,D)$ & Loss function\\
        $D_G^{*}(\mathbf{x})$ & Optimum discriminator with fixed generator $G$ \\
		\bottomrule
	\end{tabular}
\end{table}

Table \ref{table:GANs_symbols_definitions} provides the definitions for the symbols used in this section. When training GANs, it is assumed that the real data follows a probability distribution denoted as $p_X(\mathbf{x})$. The goal of GANs is to find a probability distribution that closely matches $p_X(\mathbf{x})$. The generator $G$ takes a random input vector $\mathbf{z}$ sampled from a distribution $p_Z(\mathbf{z})$, typically assumed to be a standard normal distribution but can be other distributions as well. Using this random input, $G$ produces a new sample $G(\mathbf{z})$. On the other hand, the discriminator function $D(\mathbf{x})$ acts as a classifier, assigning a probability between 0 and 1 to determine the likelihood of a sample being from the real dataset. $G$ aims to minimize the probability of the discriminator correctly classifying synthetic data, achieved by minimizing the value of $\mathbb{E}_{\mathbf{z} \sim p_Z(\mathbf{z})}[\log(1-D(G(\mathbf{z}))]$. In contrast, $D$'s objective is to correctly classify both real and synthetic data, maximizing the output probability for real samples $\mathbb{E}_{\mathbf{x} 
 \sim p_X(\mathbf{x})}[\log(D(\mathbf{x}))]$. Additionally, $D$ aims to minimize the output probability for synthetic samples $D(G(\mathbf{z}))$, accomplished by maximizing $\mathbb{E}_{\mathbf{z} \sim p_Z(\mathbf{z})}[\log(1-D(G(\mathbf{z}))]$. In other words, $D$ and $G$ engage in a two-player minimax game, with the loss function $\mathcal{L} (G,D)$ defined as follows \cite{goodfellow2014generative}:

\begin{equation} \label{eqn:GAN-minimax-loss}
 \underset{G}{\text{min}}
     \underset{D}{\text{max }} \mathcal{L} (G,D) =  \mathbb{E}_{\mathbf{x} \sim p_X(\mathbf{x})} \left[ \log(D(\mathbf{x})) \right] + \mathbb{E}_{\mathbf{z} \sim p_Z(\mathbf{z})} \left[ \log (1-D(G(\mathbf{z}))\right]
\end{equation}

With fixed $G$, the optimum discriminator $D^{*}_{G}(\mathbf{x})$ can be written as \cite{gui2021review}: 
\begin{equation} \label{eqn:GAN-optimum-discriminator}
    D^{*}_{G}(\mathbf{x}) = \frac{p_X(\mathbf{x}) }{p_X(\mathbf{x}) + p_G(\mathbf{x})}
\end{equation}
where $p_G(\mathbf{x})$ is the generator's distribution over data $\mathbf{x}$. Using Equation ({\ref{eqn:GAN-optimum-discriminator}}) with fixed $G$, Equation ({\ref{eqn:GAN-minimax-loss}}) can be written as: 
\begin{equation} \label{eqn:GAN-loss-with-optimum-D}
\begin{split}
    \underset{D}{\text{max }} \mathcal{L} (G,D) &=  \mathbb{E}_{\mathbf{x} \sim p_X(\mathbf{x})} \left[\log D^{*}_{G}(\mathbf{x})\right] + \mathbb{E}_{\mathbf{z} \sim p_Z(\mathbf{z})} \left[ \log (1- D^{*}_{G}(G(\mathbf{z}))\right] \\
       & = \mathbb{E}_{\mathbf{x} \sim p_X(\mathbf{x})} \left[\log D^{*}_{G}(\mathbf{x})\right] + \mathbb{E}_{\mathbf{x} \sim p_G(\mathbf{x})} \left[ \log (1- D^{*}_{G}(\mathbf{x}))\right] \\
       &= \mathbb{E}_{\mathbf{x} \sim p_X(\mathbf{x})} \left[\log \frac{p_X(\mathbf{x}) }{\frac{1}{2}(p_X(\mathbf{x}) + p_G(\mathbf{x}))}\right]  + \mathbb{E}_{\mathbf{x} \sim p_G(\mathbf{x})} \left[ \log \frac{p_G(\mathbf{x}) }{\frac{1}{2}(p_X(\mathbf{x}) + p_G(\mathbf{x})) }\right] -2\log 2\\
\end{split}
\end{equation}

Using the definitions of Kullback-Leibler (KL) divergence and Jensen-Shannon (JS) divergence between two probabilistic distributions $p_X(\mathbf{x})$ and $p_G(\mathbf{x})$, Equation (\ref{eqn:GAN-loss-with-optimum-D}) can be written in the form of KL and JS divergences:
\begin{equation} \label{eqn:GAN-loss-with-KL-and-JS}
\begin{split}
    \underset{D}{\text{max }} \mathcal{L} (G,D) & =           
         \mathcal{D}_{\text{KL}} \left(p_X||\frac{p_X+p_G}{2}\right)+ \mathcal{D}_{\text{KL}} \left(p_G|| \frac{p_X+p_G}{2} \right) - 2 \log 2 \\
         & = 2 \mathcal{D}_{\text{JS}} (p_X||p_G)-2 \log 2
\end{split}
\end{equation}
where $\mathcal{D}_{\text{KL}}$ and $\mathcal{D}_{\text{JS}}$ mean the KL and JS divergences, respectively.

The $D$ and $G$ are trained alternately, with one model's parameters updated while the other model's parameters are fixed. Parameters are updated using the feedback from the loss function in Equation (\ref{eqn:GAN-minimax-loss}). This continues until reaching the equilibrium state.

%%%%%%%%%%%%%%%%%%%%%%%%%%%%%%%%%%%%%%%%
\subsection{A brief theory of NFs}

NFs make use of invertible functions, also known as bijective functions, along with the change of variables rule, to transform the intricate distribution of the data into a simple distribution. This process has two directions: training and generation. During training, the model learns to convert the complex input data into a simpler distribution, usually a normal distribution. In the generation phase, the model uses the learned transformations in reverse to create new data samples that follow the complex distribution of the original training data. In this subsection, we will briefly introduce the theory of NFs. Table \ref{table:NFs_symbols_definitions} lists the definitions of all the symbols used in this section. 

\begin{table}[htbp] 
	\footnotesize
	\captionsetup{justification=centering}
	\caption{Definitions of mathematical symbols for NFs.}
	\label{table:NFs_symbols_definitions}
	\centering
	\begin{tabular}{cl}
		\toprule
	    Symbols & Meanings \\ 
		\midrule
		$\mathbf{z}$  & Random input vector  \\
		$\mathbf{x}$  & Samples from input training data    \\
		$p_Z (\mathbf{z})$ &  Distribution of $\mathbf{z}$\\
	    $p_X (\mathbf{x})$ &  Distribution of $\mathbf{x}$\\
        $g$  & Bijector function mapping from $\mathbf{z}$ to $\mathbf{x}$\\
        $f$  & Bijector function mapping from $\mathbf{x}$ to $\mathbf{z}$\\
        $D$   &  Dimension of input variables $\mathbf{x}$\\
        $d$   & Number of $\mathbf{x}$ samples that were not updated in this layer. \\
        $s$  & Scaling function\\
        $t$  & Shifting function\\
		\bottomrule
	\end{tabular}
\end{table}

Let's consider a continuous latent variable $Z$, whose samples $\mathbf{z}$ follow a simple normal distribution $p_{Z}(\mathbf{z})$. The main concept behind NFs is to transform this initial distribution $p_{Z}(\mathbf{z})$ into a more complex distribution by applying a sequence of invertible transformations using a bijective function $g$. A bijective function creates a one-to-one invertible mapping between datasets. The purpose of this bijective function $g$ is to map samples $\mathbf{z}$ from the latent distribution to generate samples $\mathbf{x}$ that approximate the distribution of the training data.

During the training process, we need to map the input data samples $\mathbf{x}$, following the distribution $p_{X}(\mathbf{x})$, to their corresponding latent samples $\mathbf{z}$. This mapping is performed using the inverse bijective function $f = g^{-1}$. Each sample in $\mathbf{x}$ is mapped to a unique sample in $\mathbf{z}$ \cite{kobyzev2020normalizing}. By applying the change of variables formula, we can obtain the expression for the distribution $p_{X}(\mathbf{x})$ as follows:
\begin{equation}\label{eqn:NFs-probability-dis-x}
    p_{X}(\mathbf{x}) = p_{Z}(f(\mathbf{x})) \left| \det\left( \frac{df(\mathbf{x})}{d \mathbf{x}^\top}\right) \right|
\end{equation}

Now the maximum likelihood objective can be written as:
\begin{equation}\label{eqn:NFs_loss}
    \log (p_{X}(\mathbf{x})) = \log (p_{Z}(f(\mathbf{x}))) + \log \left( \left| \det \left(  \frac{df(\mathbf{x})}{d \mathbf{x}^\top} \right) \right|  \right)
\end{equation}
where $\left| \det \left( \frac{df(\mathbf{x})}{d \mathbf{x}^\top} \right) \right|$ is the determinant of the Jacobian of $f$. Here, $f$ is a composition of sequence of invertible functions. Since each bijective function $f$ is represented by a DNN, during training, the DNN parameters are optimized to maximize the log-likelihood.

\begin{figure}[!ht]
    \centering
	\captionsetup{justification=centering}
    \includegraphics[width=0.9\textwidth]{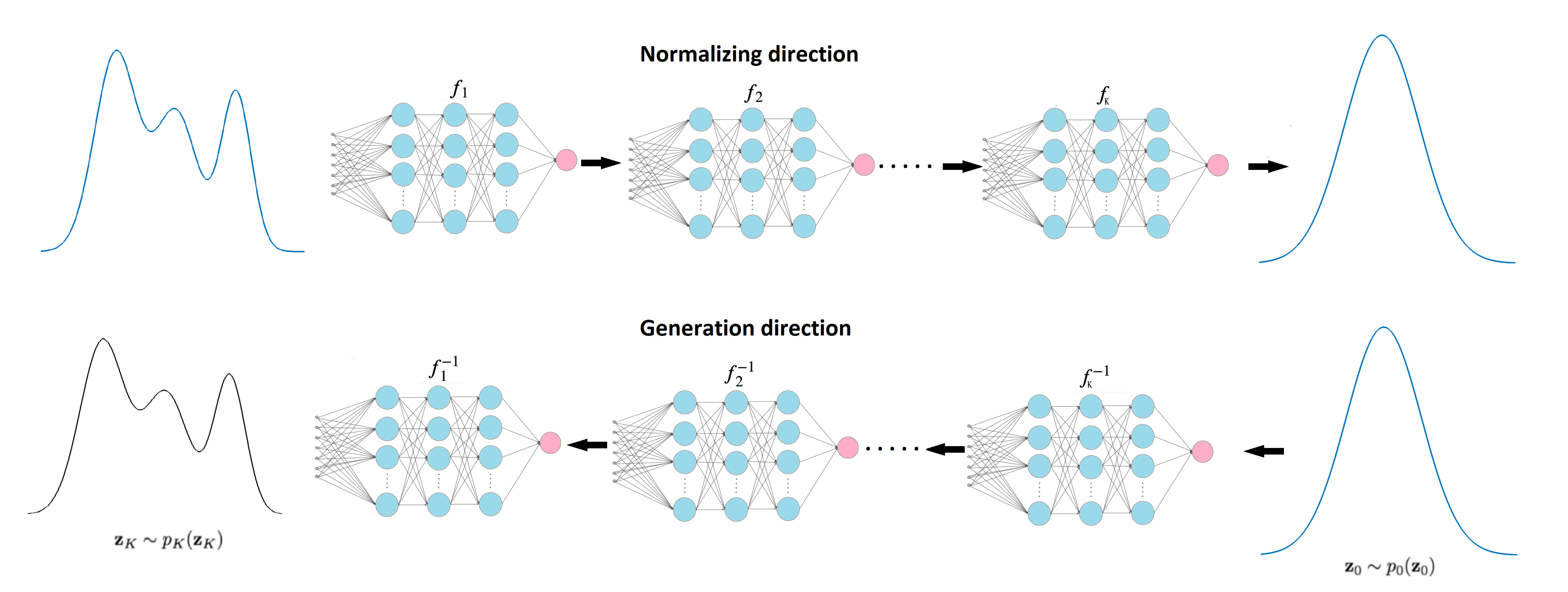}
    \caption{Illustration of the training (normalizing) and generation directions of NFs.}
    \label{fig:NF-architecture}
\end{figure}

In order to generate samples of $\mathbf{x}$, it is necessary to invert the transformations. We start by sampling a value $\mathbf{z}$ from the latent space and then apply the inverse bijective function $\mathbf{x} = f^{-1}(\mathbf{z}) = g(\mathbf{z})$. By using Equation (\ref{eqn:NFs-probability-dis-x}), we can calculate the probability density for $\mathbf{x}$. Figure \ref{fig:NF-architecture} illustrates the two directions: training (normalizing) and generation.

It is necessary to ensure efficient computation of the determinant of the Jacobian, and that the function $f$ can be easily inverted. One approach to address this issue is the utilization of multi-scale architecture methods, which involve dividing the inputs into two blocks and using components that can be readily inverted. Among these methods is the real-valued non-volume preserving (real NVP) method \cite{dinh2016density}, which offers efficient evaluation of both $f$ and $f^{-1}$, along with a straightforward and effective computation of the Jacobian determinant.

In real NVP networks, the bijections employed are known as coupling layers, specifically affine coupling layers. Each layer applies scaling and shifting operations to a portion of the data, while the other part is utilized to compute the scaling and shifting functions $s$ and $t$ respectively. The bijective function $f$ is constructed by stacking a series of such bijections. In an affine layer with $D$-dimensional input variables $\mathbf{x}$ and $\mathbf{z}$ representing the layer's output, the operations are performed as follows:
\begin{equation} \label{eqn:NF-generation-pass-of-real-nvp}
\begin{split}
    \mathbf{z}_{1:d} =  &\mathbf{x}_{1:d}\\
    \mathbf{z}_{d+1:D} &= \mathbf{x}_{d+1:D}\odot \exp(s(\mathbf{x}_{1:d}) ) + t(\mathbf{x}_{1:d})
\end{split}
\end{equation}

This bijection is part of the bijector function $f$, so it maps from $\mathbf{x}$ to $\mathbf{z}$. Where $\odot$ is the element wise product, $d<D$ and $s$ and $t$ are functions that map from $R^d$ to $R^{D-d}$. $s$ and $t$ are represented by DNNs. In order to go from $\mathbf{z}$ to $\mathbf{x}$, this process is reversed: 
\begin{equation} \label{eqn:NF-training-pass-of-real-nvp}
\begin{split}
    \mathbf{x}_{1:d} = &\mathbf{z}_{1:d}\\
    \mathbf{x}_{d+1:D} &= (\mathbf{z}_{d+1:D} - t(\mathbf{z}_{1:d})) \odot \exp( -s(\mathbf{z}_{1:d}))
\end{split}
\end{equation}

The Jacobian in this case is a triangular matrix. Its determinant will become the product of the diagonals, thus, the logarithm of the Jacobian determinant simplifies to $\sum_{i} s_{i}(x_{1:d})$.

%%%%%%%%%%%%%%%%%%%%%%%%%%%%%%%%%%%%%%%%
\subsection{A brief theory of VAEs}

The VAEs model is a generative model that utilizes autoencoders. It consists of three main components: the encoder, the latent space (also known as the bottleneck), and the decoder. The objective of VAEs is to learn or approximate the distribution of the training data in order to generate new samples. The key distinction between conventional autoencoders and VAEs lies in the encoding process of the input. In autoencoders, the encoder takes the input data and encodes it into the latent space, ensuring that only the most important features are retained and can be subsequently restored. The decoder then utilizes the encoded information to reconstruct the input by reversing the encoding process, effectively returning from the latent space to the initial space. This process allows for tasks such as image reconstruction \cite{han2022inference}, denoising \cite{Liu_2020_CVPR_Workshops} and anomaly detection \cite{GONZALEZMUNIZ2022108065} \cite{CHATTERJEE2022106093}. However, autoencoders are not suitable for data generation as they are trained to encode and decode with minimal loss, without considering the organization of the latent space. Therefore, if we were to disregard the encoder and solely rely on the decoder as a generator, it would be challenging to determine appropriate values to assign to the latent space during the generation phase. If a random vector is used as an input, the resulting output would lack meaningful interpretation.

\begin{figure}[!ht]
    \centering
	\captionsetup{justification=centering}
    \includegraphics[width=0.7\textwidth]{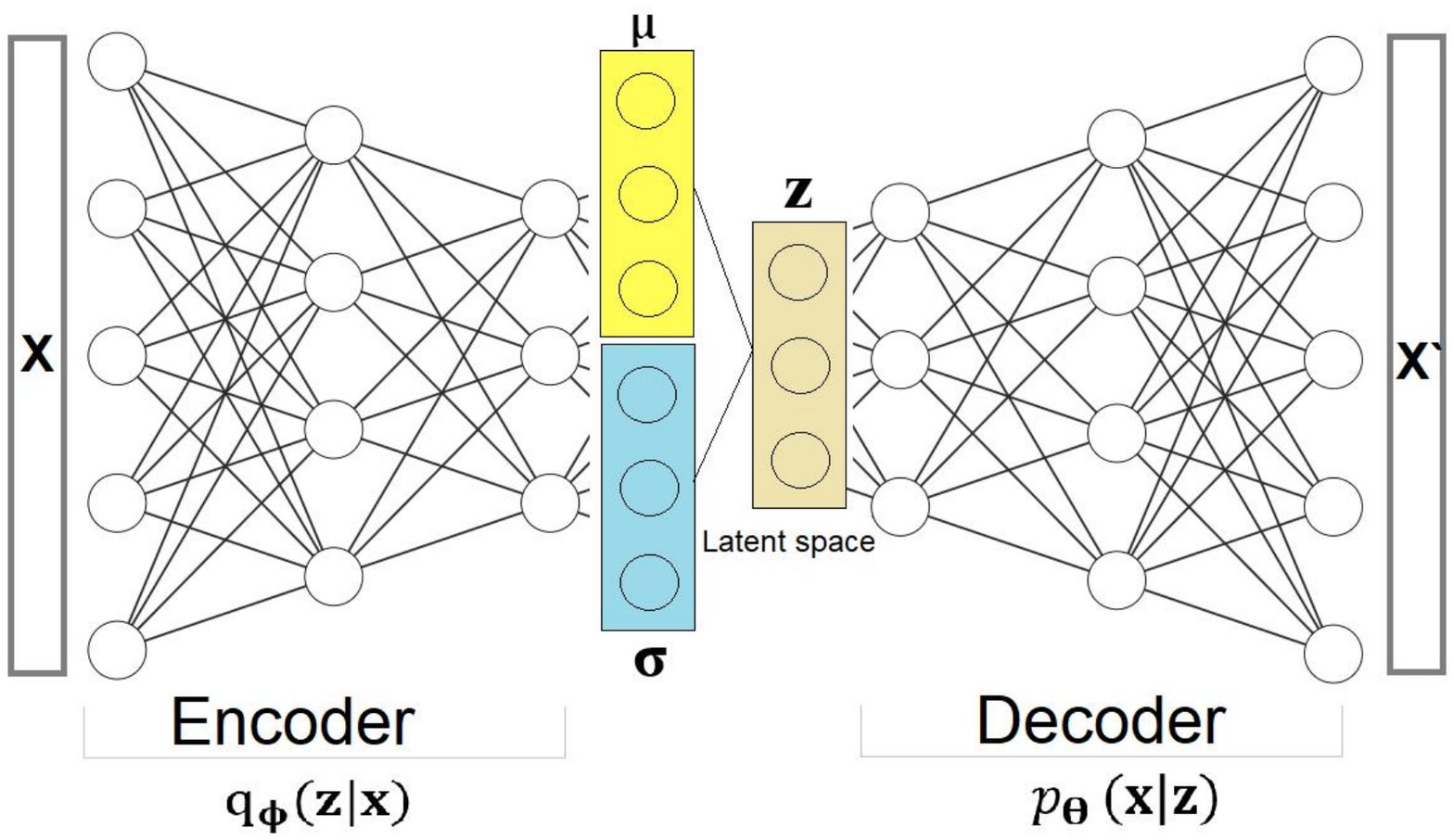}
    \caption{Illustration of the typical structure of VAEs.}
    \label{fig:VAE-structure}
\end{figure}

The concept behind VAEs is to encode the input data into a distribution instead of a fixed vector \cite{kingma2013auto}. As a result, the latent space of VAEs consists of two vectors that represent the mean value and standard deviation of the encoded (multi-dimensional) distribution, as shown in Figure \ref{fig:VAE-structure}. This approach creates a structured latent space that can be effectively utilized for data generation. To generate new samples, the VAEs model takes a random input vector sampled from the encoded distributions in the latent space and feeds it into the decoder network. The decoder network then produces the output, generating new samples $\mathbf{x}$ based on the input vector.

Define $\mathbf{x}$ as a set of observed variables with a distribution denoted as $p(\mathbf{x})$ and $\mathbf{z}$ as a set of latent variables with a joint distribution $p(\mathbf{z},\mathbf{x})$. By applying Bayes' rule, we can compute the conditional probability $p(\mathbf{z}|\mathbf{x})$ as:

\begin{equation}\label{eqn:Bayes-rule}
    p(\mathbf{z}|\mathbf{x}) = \frac{p(\mathbf{x}|\mathbf{z}) p(\mathbf{z})}{p(\mathbf{x})}
\end{equation}

The computation of $p(\mathbf{z}|\mathbf{x})$ poses a challenge due to the intractable integral $p(\mathbf{x}) = \int p(\mathbf{z}|\mathbf{x}) p(\mathbf{z}) d\mathbf{z}$. As an alternative approach, $p(\mathbf{z}|\mathbf{x})$ can be approximated by another distribution $q(\mathbf{z}|\mathbf{x})$ that has a tractable solution. This is done using variational inference (VI) \cite{blei2017variational}. The core idea behind VI is to reformulate the inference problem as an optimization problem, where $p(\mathbf{z}|\mathbf{x})$ is approximated by the simpler distribution $q(\mathbf{z}|\mathbf{x})$, often chosen to be Gaussian in practice, and then minimize the discrepancy between them. This discrepancy can be measured using the KL divergence, which quantifies the difference between two probability distributions as follows:

\begin{align} \label{eqn:KL-divergence}
    \mathcal{D}_{\text{KL}} ( q_{\bm{\phi}}(\mathbf{z}|\mathbf{x}) || p_{\bm{\theta}}(\mathbf{z}|\mathbf{x}) ) &= \sum_\mathbf{z} q_{\bm{\phi}}(\mathbf{z}|\mathbf{x}) \log \left( \frac{ q_{\bm{\phi}}(\mathbf{z}|\mathbf{x}) }{ p_{\bm{\theta}}(\mathbf{z}|\mathbf{x})} \right ) \\
    &= \mathbb{E}_{\mathbf{z} \sim q_{\bm{\phi}}(\mathbf{z}|\mathbf{x}) } \left[ \log(q_{\bm{\phi}}(\mathbf{z}|\mathbf{x}) ) - \log( p_{\bm{\theta}}(\mathbf{z}|\mathbf{x}) ) \right]
\end{align}

Here, the encoder network is represented by $q_{\bm{\phi}}(\mathbf{z}|\mathbf{x})$, and the decoder network is represented by $p_{\bm{\theta}}(\mathbf{z}|\mathbf{x})$. Table \ref{table:VAEs_symbols_definitions} provides a comprehensive list of the symbols used in this section along with their definitions. The parameters $\bm{\phi}$ and $\bm{\theta}$ correspond to the network parameters that will be optimized during the VAEs model training. By substituting Equation (\ref{eqn:Bayes-rule}) into Equation (\ref{eqn:KL-divergence}), simplifying and rearranging, we obtain the following expression for the loss function:

\begin{equation} \label{eqn:loss-function-derivation-vae}
\begin{aligned}
    \mathcal{L} (\bm{\theta},\bm{\phi})&= - \mathbb{E}_{\mathbf{z} \sim q_{\bm{\phi}}(\mathbf{z}|\mathbf{x}) } \left[ \log(  p_{\bm{\theta}}(\mathbf{x}|\mathbf{z}) )\right] + \mathcal{D}_{\text{KL}} ( q_{\bm{\phi}}(\mathbf{z}|\mathbf{x}) || p_{\bm{\theta}}(\mathbf{z}) ) \\
    &=  -\log ( p_{\bm{\theta}}(\mathbf{x}) ) + \mathcal{D}_{\text{KL}} ( q_{\bm{\phi}}(\mathbf{z}|\mathbf{x}) || p_{\bm{\theta}}(\mathbf{z}|\mathbf{x}) )
\end{aligned}
\end{equation}

\begin{table}[htbp] 
	\footnotesize
	\captionsetup{justification=centering}
	\caption{Definitions of mathematical symbols for VAEs and CVAEs.}
	\label{table:VAEs_symbols_definitions}
	\centering
	\begin{tabular}{cl}
		\toprule
	    Symbols & Meanings \\ 
		\midrule
		$\mathbf{z}$  & Latent variables  \\
		$\mathbf{x}$  & Samples from real dataset    \\
        $\bm{\phi}, \bm{\theta}$ & Neural network parameters for the encoder and decoder \\
		$p_{\bm{\theta}} (\mathbf{z})$ &  Distribution of $\mathbf{z}$\\
	    $p_{\bm{\theta}} (\mathbf{x})$ &  Distribution of $\mathbf{x}$\\
        $p_{\bm{\theta}} (\mathbf{x}|\mathbf{z})$ & Decoder network \\ 
        $q_{\bm{\phi}} (\mathbf{z}|\mathbf{x})$ & Encoder network\\
        $\mathcal{D}_{\text{KL}}$ & The Kullback–Leibler divergence\\
        $\mathcal{L} (\bm{\theta},\bm{\phi}) $ &The ELBO function\\ 
		\bottomrule
	\end{tabular}
\end{table}

In Equation (\ref{eqn:loss-function-derivation-vae}), the KL divergence term serves as a regularization term, ensuring that the learned distributions closely match the assumed Gaussian prior. The expectation term represents the reconstruction loss, measuring the model's ability to accurately reconstruct the input data. The loss function $ \mathcal{L} (\bm{\theta},\bm{\phi})$ is referred to as the evidence lower bound (ELBO). Lower bound comes from the fact that KL divergence is always non-negative. Therefore, in the training of VAEs, minimizing the loss, corresponds to maximizing the lower bound of the probability for generating real data samples.

%%%%%%%%%%%%%%%%%%%%%%%%%%%%%%%%%%%%%%%%
\subsection{A brief theory of CVAEs}

One of the disadvantages of the VAEs model is that it cannot generate specific data. For example, when using the MNIST handwritten digits dataset, it cannot generate samples for a specific number only, such as the number one. Instead, it will generate random numbers from zero to nine. To address this limitation, CVAEs can be used. The CVAEs model works similarly to the VAEs model but incorporates additional data for conditioning. During the training process, instead of using the training data alone, the CVAEs utilize labels along with the data for training both the encoder and the decoder. By including the labels, the decoder can learn to generate specific data corresponding to the given labels.

Once the training is completed, the decoder can take a random vector from the latent space and combine it with a label to select specific data to be generated. This means that providing different labels for the same input in the latent space can result in different outputs. Therefore, CVAEs allow for generating targeted data based on the specified labels. This can be expressed mathematically in the loss function by conditioning the encoder, denoted as $q_{\bm{\phi}}(\mathbf{z}|\mathbf{x})$, and the decoder, denoted as $p_{\bm{\theta}}(\mathbf{x}|\mathbf{z})$, networks on the label $c$. By modifying the original VAEs loss function from Equation (\ref{eqn:loss-function-derivation-vae}), the loss function of CVAEs can be expressed as follows:

\begin{equation} \label{eqn:loss-function-cvae}
    \mathcal{L} (\bm{\theta},\bm{\phi}) = - \mathbb{E}_{\mathbf{z} \sim q_{\bm{\phi}}(\mathbf{z}|\mathbf{x},c) } \left[ \log(  p_{\bm{\theta}}(\mathbf{x}|\mathbf{z},c) )\right] + \mathcal{D}_{\text{KL}} ( q_{\bm{\phi}}(\mathbf{z}|\mathbf{x},c) || p_{\bm{\theta}}(\mathbf{z}|c) )
\end{equation}

The CVAEs model is trained by optimizing the network parameters $\bm{\theta}$ and $\bm{\phi}$ to minimize the loss function. This minimization process corresponds to maximizing the lower bound of the probability for generating real data samples. Through the training, CVAEs learn to better encode and decode the latent space, capturing meaningful representations of the input data and generating samples that align with the specified label.

%%%%%%%%%%%%%%%%%%%%%%%%%%%%%%%%%%%%%%%%%%%%%%%%%%%%%%%%%%%%%%%%%%%%%%%%%%%%%%%%%%%%%%%%
%%%%%%%%%%%%%%%%%%%%%%%%%%%%%%%%%%%%%%%%%%%%%%%%%%%%%%%%%%%%%%%%%%%%%%%%%%%%%%%%%%%%%%%%
\section{Results}
\label{sec:Results}

This section presents the results for all the DGMs and comparison of their performance in generating samples that are most similar to the TRACE simulations. All models were trained using the same dataset consisting of 200 samples. Each sample in the dataset includes the five PMPs ($\texttt{P1008}$, $\texttt{P1012}$, $\texttt{P1022}$, $\texttt{P1028}$, and $\texttt{P1029}$) and the four void fraction values ($\texttt{VoidF1}$, $\texttt{VoidF2}$, $\texttt{VoidF3}$, and $\texttt{VoidF4}$). The training dataset was standardized before being used to train the DGMs. To evaluate the performance of the models, each model was utilized to generate 500 synthetic samples. Following the generation step, the range and correlation of the PMPs were examined. 
%As the PMPs in the training data were randomly generated within the range of $(0.0,5.0)$ without any correlations, it is important to assess the randomness and correlations in the generated samples for each model.
Samples falling within the desired range were selected for the validation step. For validation, the corresponding ``real'' void fraction data was obtained by running the TRACE code at the same inputs (five PMPs) of the synthetic samples. A comparison was then made between the void fractions obtained from TRACE and the DGM models to evaluate the performance of the DGMs.

\begin{figure}[ht!]
	\captionsetup{justification=centering}
	\centering
	\begin{subfigure}{0.495\textwidth}
		\centering
		\includegraphics[width=0.95\linewidth]{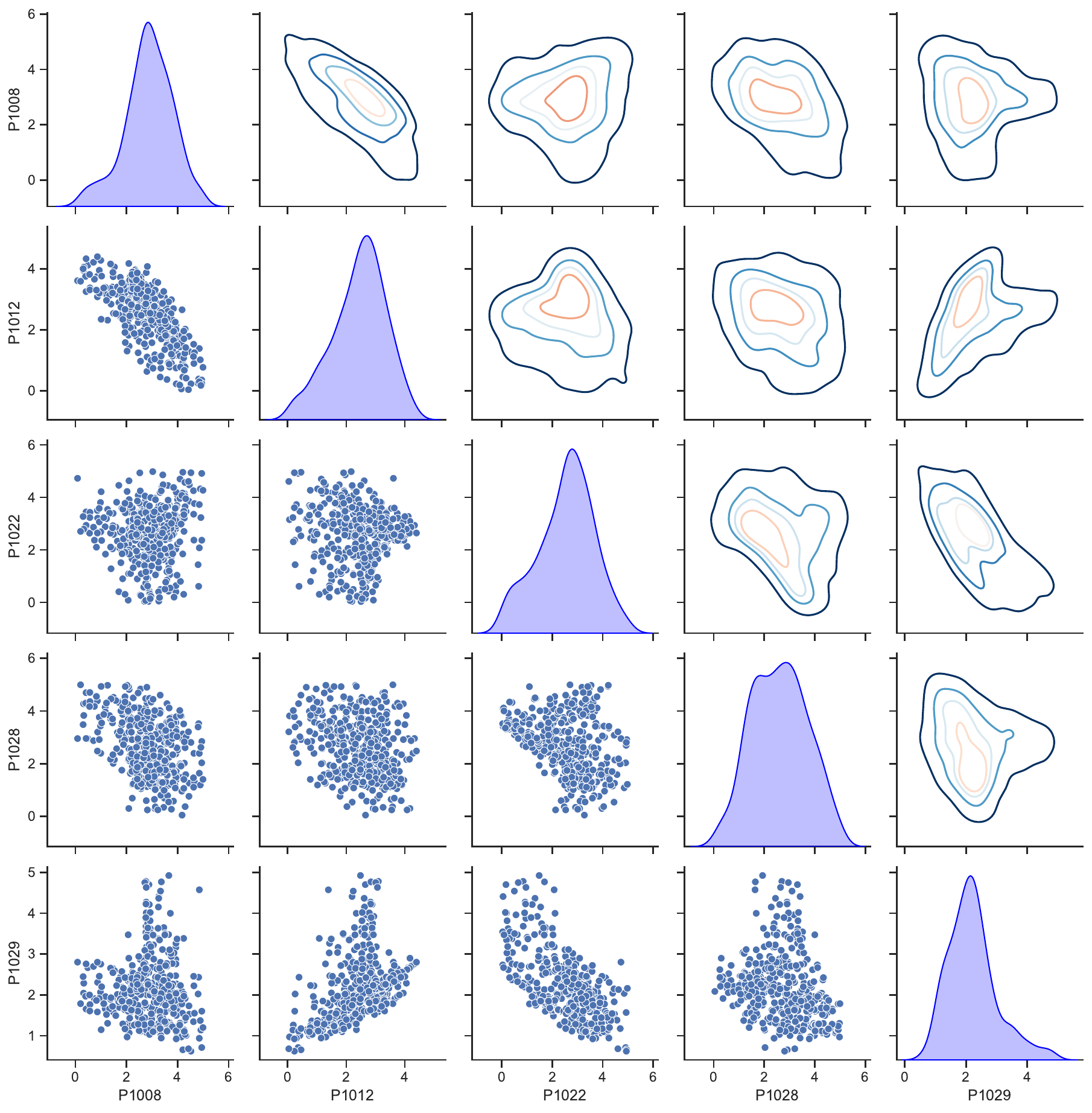}
		\caption{GANs}
		\label{fig:GAN-samples-dist}
	\end{subfigure}	
	\begin{subfigure}{0.495\textwidth}
		\centering
		\includegraphics[width=0.95\linewidth]{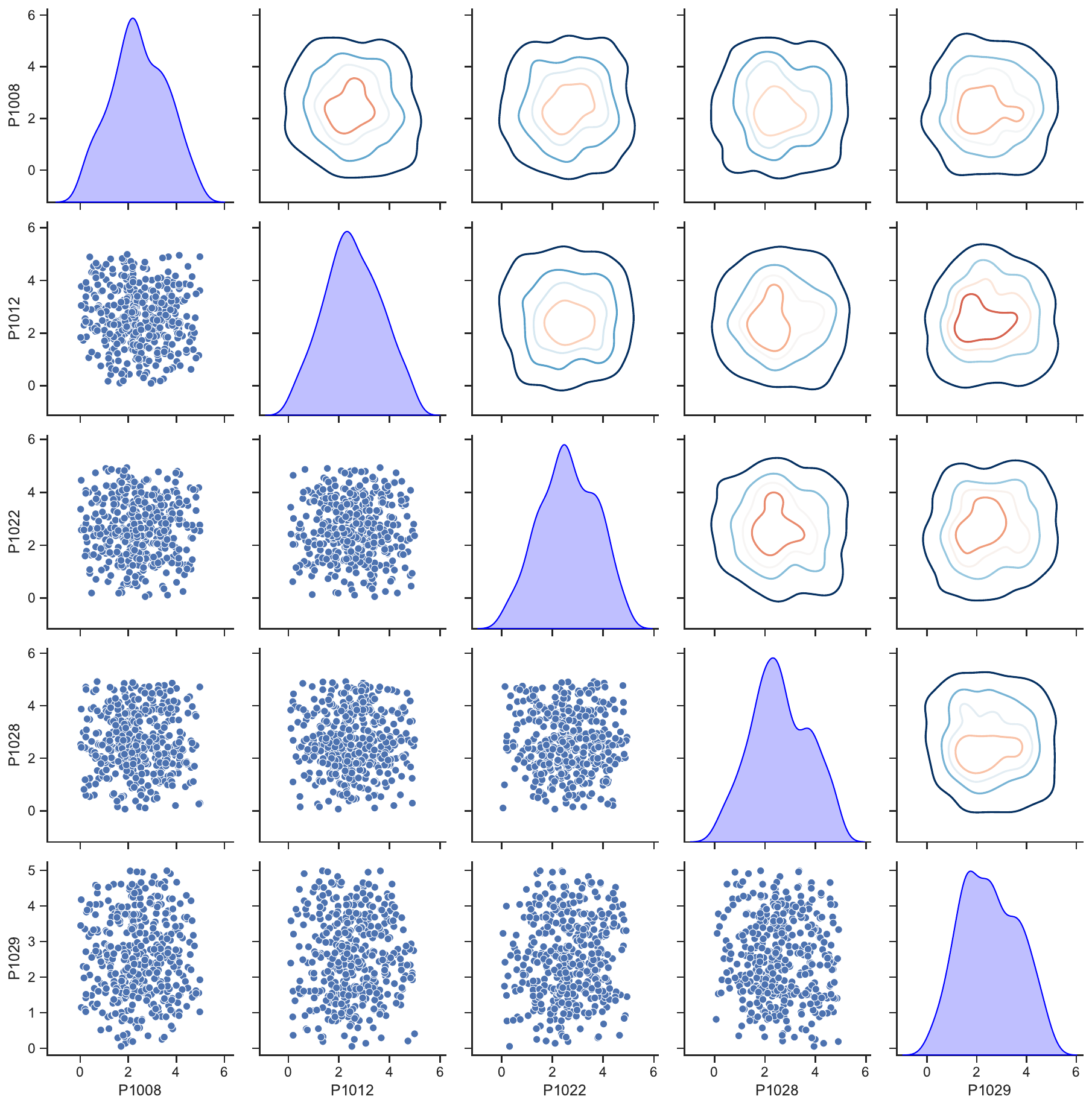}
		\caption{NFs}
		\label{fig:NF-samples-dist}
	\end{subfigure}	
	\begin{subfigure}{0.495\textwidth}
		\centering
		\includegraphics[width=0.95\linewidth]{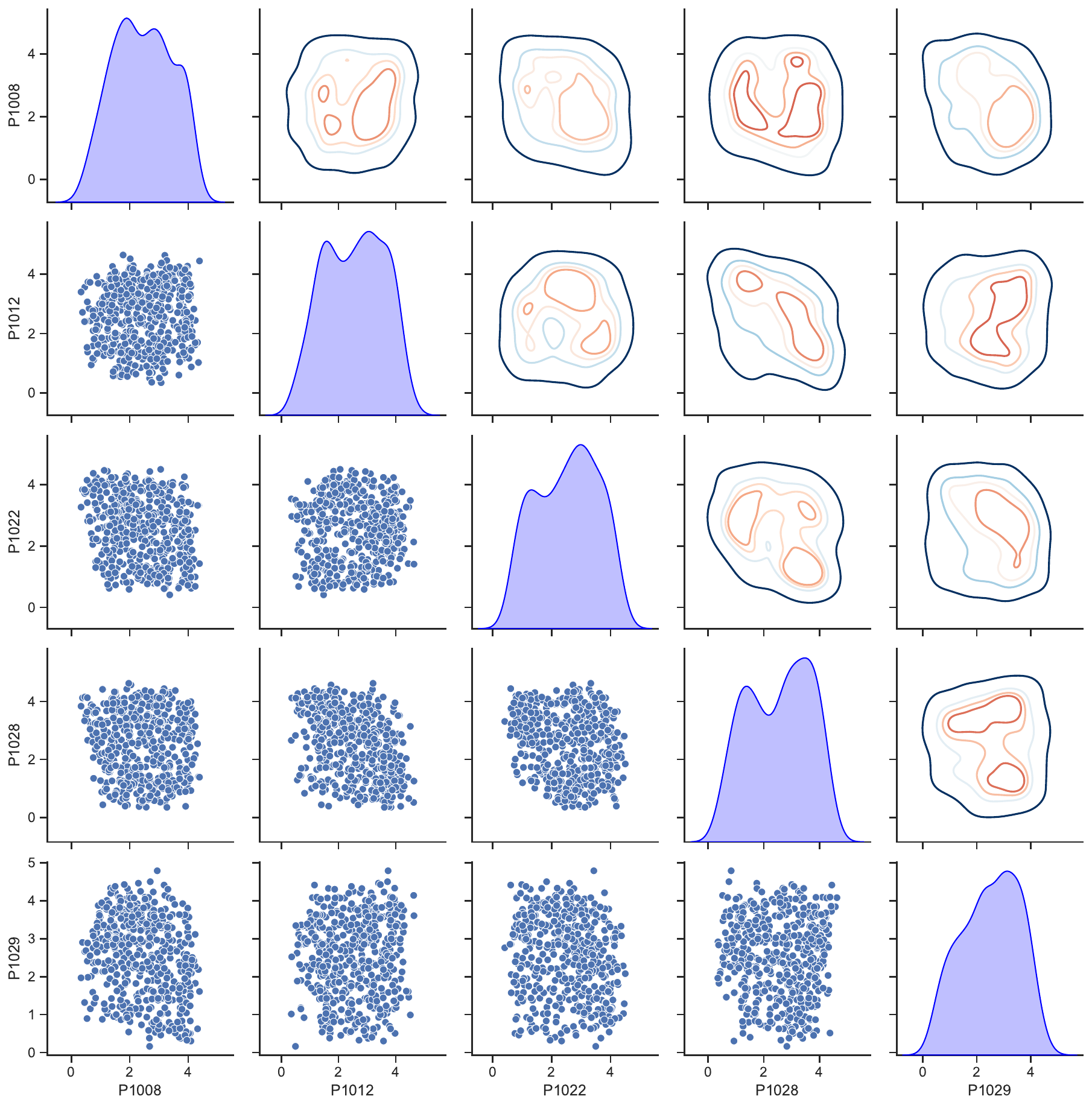}
		\caption{VAEs}
		\label{fig:VAE-samples-dist}
	\end{subfigure}	
	\begin{subfigure}{0.495\textwidth}
		\centering
		\includegraphics[width=0.95\linewidth]{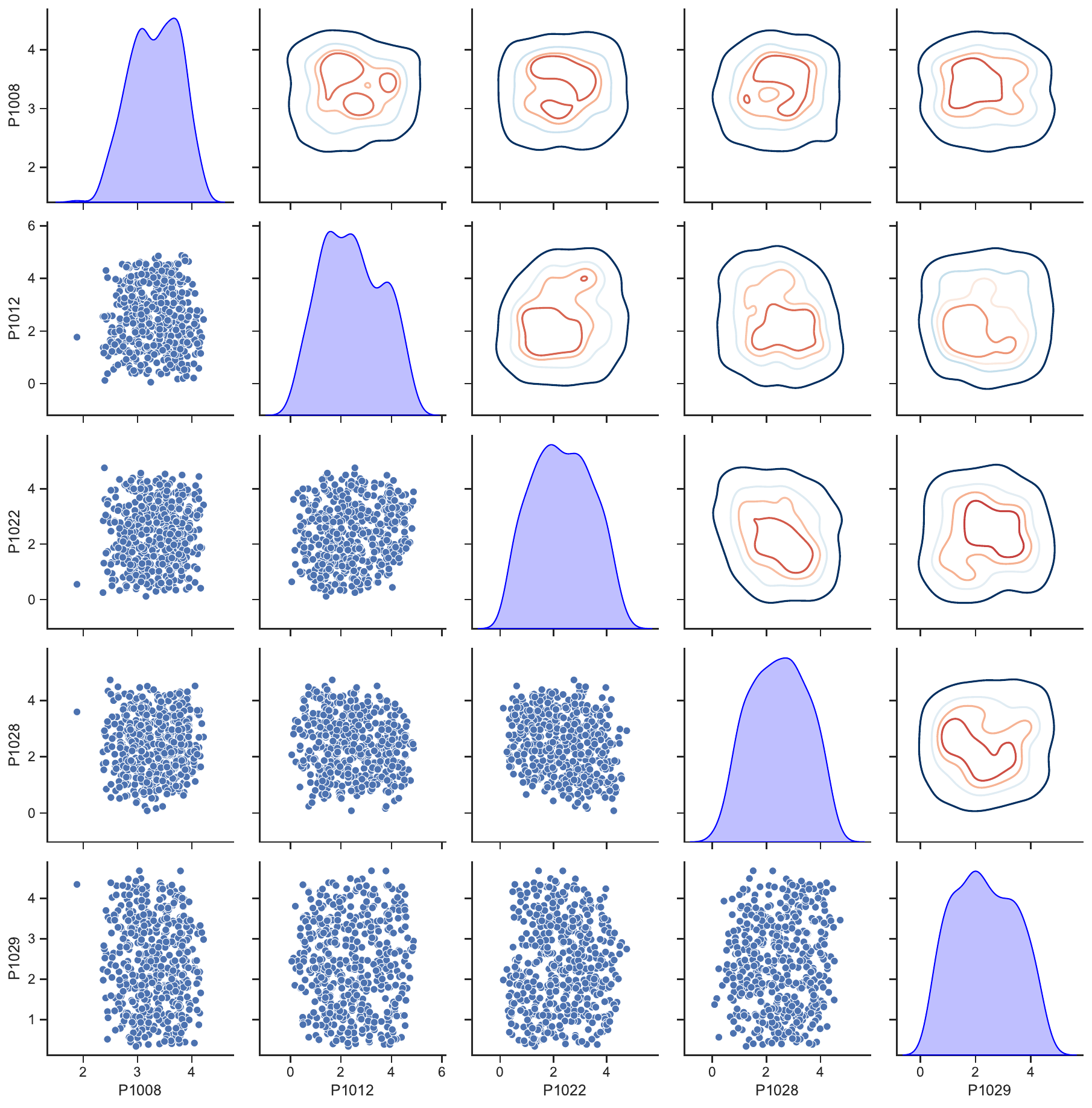}
		\caption{CVAEs}
		\label{fig:CVAE-samples-dist}
	\end{subfigure}	
	\caption{Distributions of the PMP samples generated by the four DGMs, after removing the samples that are out of the training domain. The diagonal plots are the marginal distributions for each PMP. The lower triangle plots are the pair-wise scatter plots of the sample, while the upper triangle plots are the pair-wise probability density contours.}
	\label{fig:Generated_sample_distribution}
\end{figure}

\begin{figure}[htbp]
	\captionsetup{justification=centering}
    \centering
	\begin{subfigure}{\textwidth}
		\centering
		\includegraphics[width=0.95\linewidth]{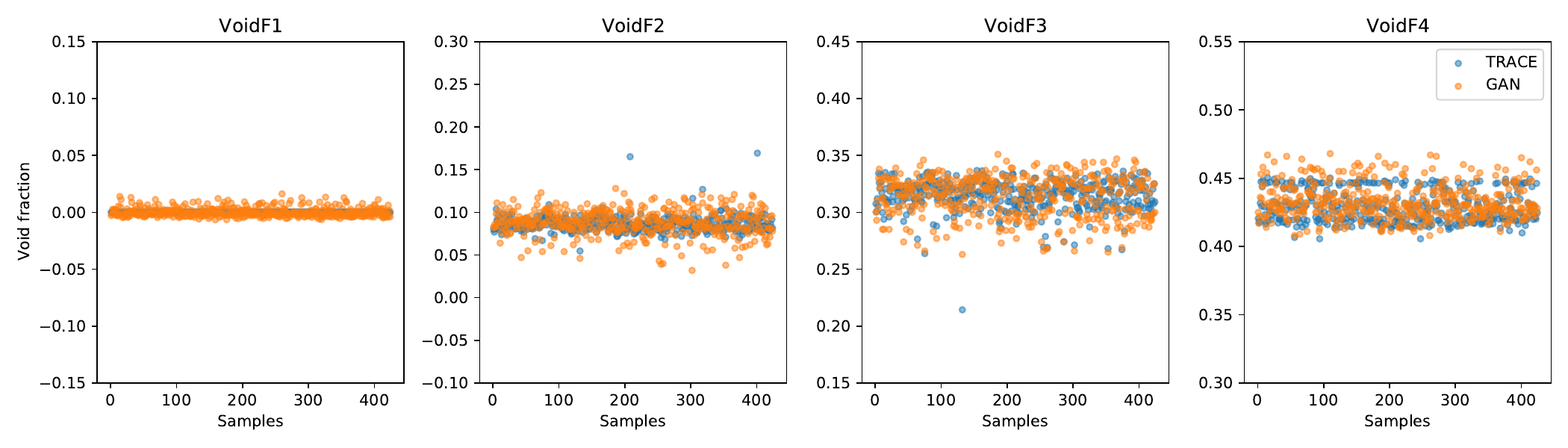}
		\caption{GANs}
		\label{fig:GAN-predictions}
	\end{subfigure}
	\begin{subfigure}{\textwidth}
		\centering
		\includegraphics[width=0.95\linewidth]{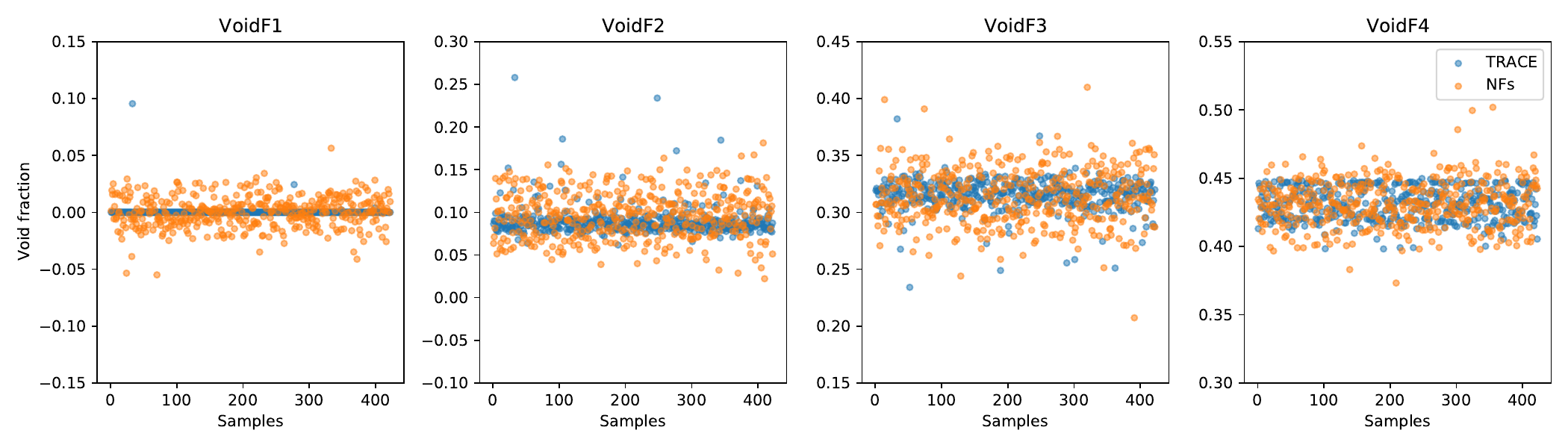}
		\caption{NFs}
		\label{fig:NF-predictions}
	\end{subfigure}
	\begin{subfigure}{\textwidth}
		\centering
		\includegraphics[width=0.95\linewidth]{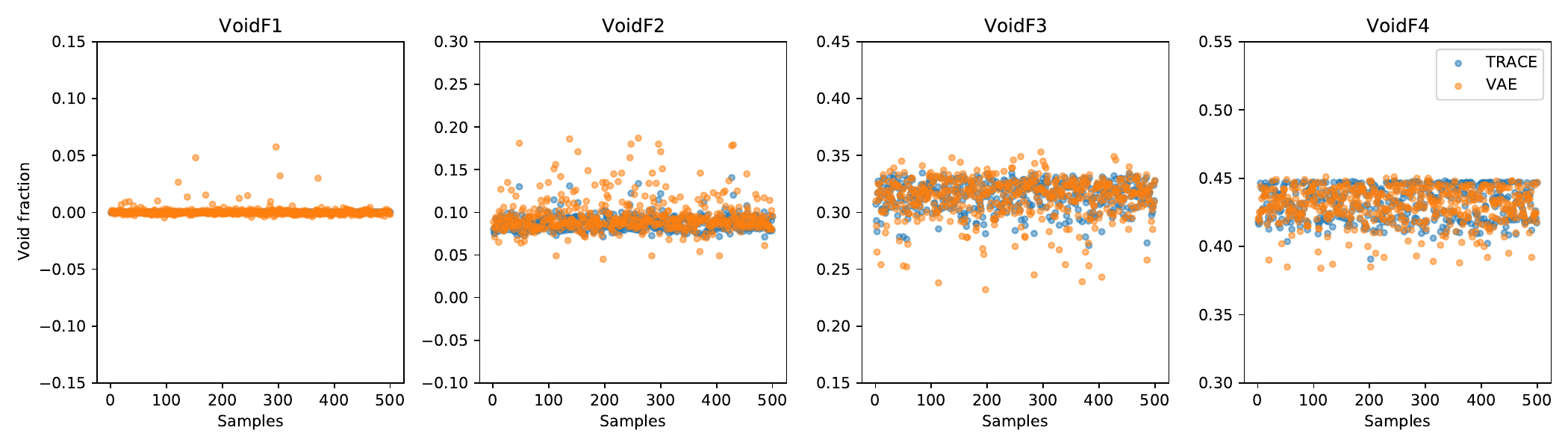}
		\caption{VAEs}
		\label{fig:VAE-predictions}
	\end{subfigure}
	\begin{subfigure}{\textwidth}
		\centering
		\includegraphics[width=0.95\linewidth]{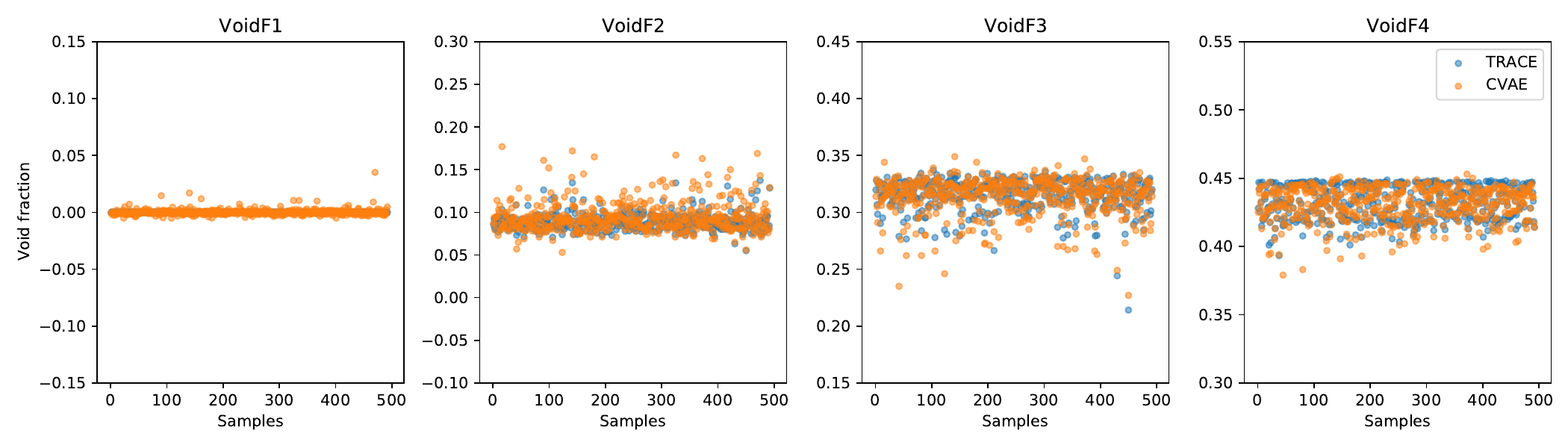}
		\caption{CVAEs}
		\label{fig:CVAE-predictions}
	\end{subfigure}	
	\caption{Comparison of the DGM-generated and TRACE-simulated void fractions.}
	\label{fig:predictions-comparison}
\end{figure}

\begin{figure}[ht!]
	\captionsetup{justification=centering}
	\centering
	\begin{subfigure}{0.495\textwidth}
		\centering
		\includegraphics[width=0.99\linewidth]{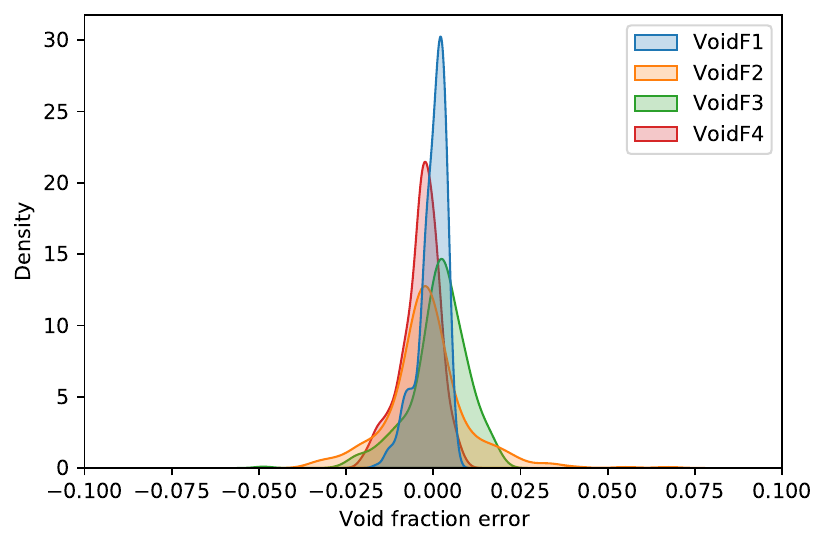}
		\caption{GANs}
		\label{fig:GAN-error-distribution}
	\end{subfigure}
	\begin{subfigure}{0.495\textwidth}
		\centering
		\includegraphics[width=0.975\linewidth]{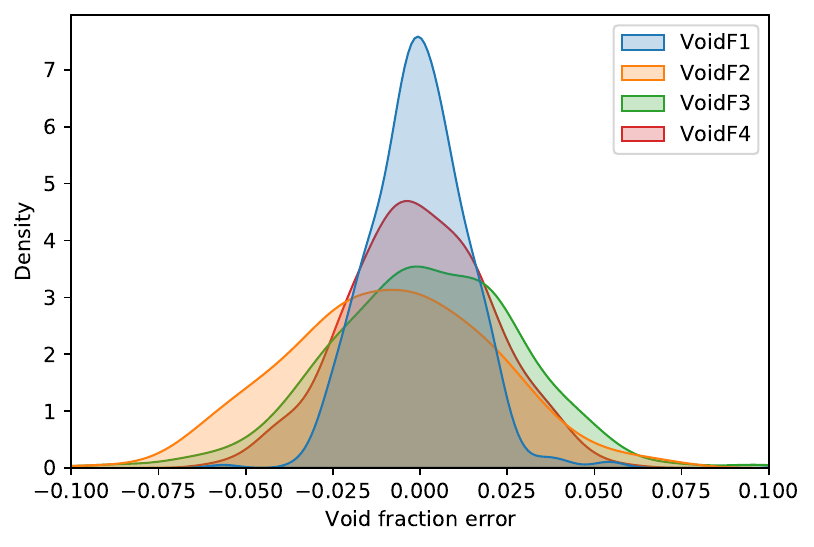}
		\caption{NFs}
		\label{fig:NF-error-distribution}
	\end{subfigure}
	\begin{subfigure}{0.495\textwidth}
		\centering
		\includegraphics[width=0.99\linewidth]{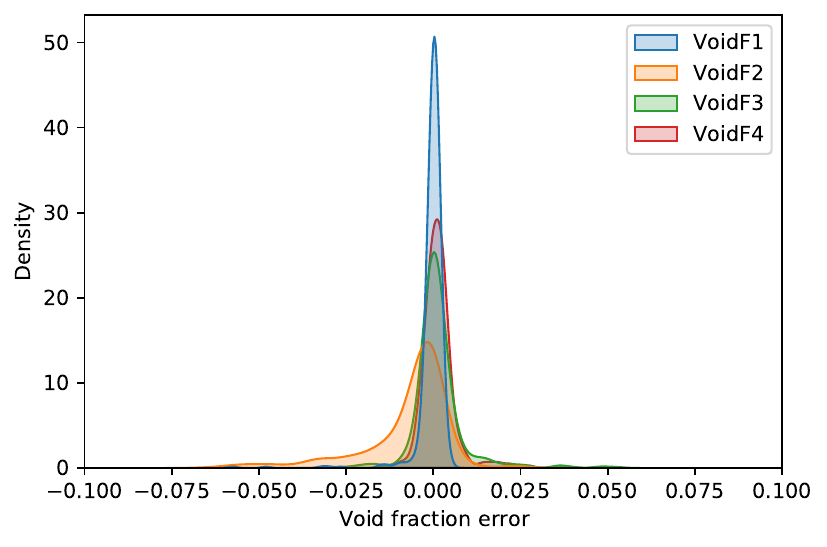}
		\caption{VAEs}
		\label{fig:VAE-error-distribution}
	\end{subfigure}
	\begin{subfigure}{0.495\textwidth}
		\centering
		\includegraphics[width=0.99\linewidth]{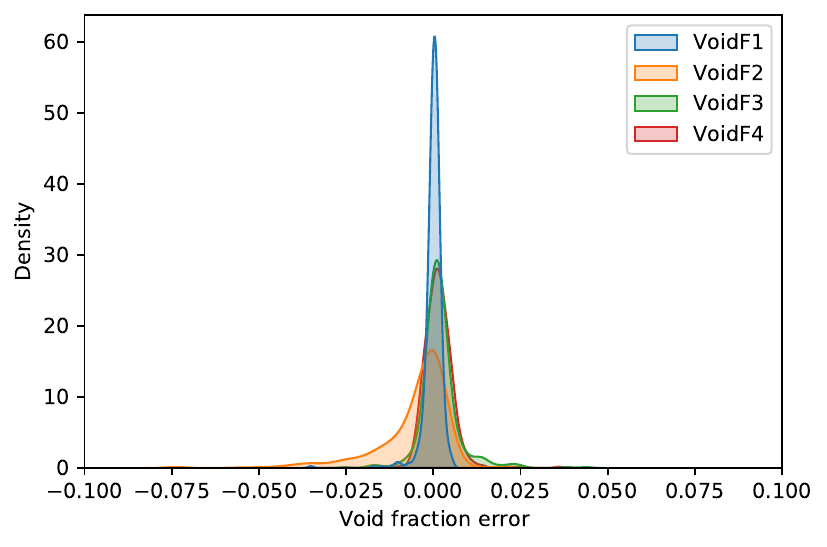}
		\caption{CVAEs}
		\label{fig:CVAE-error-distribution}
	\end{subfigure}	
	\caption{Distributions of the errors between DGM-generated and TRACE-simulated void fractions. The void fractions in the x axes are unitless, i.e., a value of 0.10 means a void fraction of 10\%.}
	\label{fig:error-distributions-comparison}
\end{figure}

To get a direct comparison, we will first present all the results. Figure \ref{fig:Generated_sample_distribution} shows the samples generated by the four DGMs, along with the marginal and pair-wise distributions. Figure \ref{fig:predictions-comparison} presents direct comparisons of the DGM-generated and TRACE-simulated void fractions. Finally, the distributions of the errors between DGM-generated and TRACE-simulated void fractions are included in Figure \ref{fig:error-distributions-comparison}. Sections \ref{sec:Results-GAN} - \ref{sec:Results-CVAE} provide details of the implementation of each DGM, followed by comparison and discussions in Section \ref{sec:Results-comparison}.

%%%%%%%%%%%%%%%%%%%%%%%%%%%%%%%%%%%%%%%%
\subsection{GANs results}
\label{sec:Results-GAN}

The GANs model utilized two DNNs as the generator and the discriminator, respectively. Each DNN consists of two hidden layers. The GANs model was trained for 30,000 epochs. The activation and loss functions for the generator and the discriminator networks are listed in Table \ref{table:Activation_and_Loss_functions}. The training process resulted in a discriminator accuracy close to $50\%$ for both real and synthetic samples, which is desirable.

\begin{table}[htbp]
	\footnotesize
	\captionsetup{justification=centering}
	\caption{Activation and loss functions used in GANs.}
	\label{table:Activation_and_Loss_functions}
	\centering
	\begin{tabular}{lll}
	    \toprule
        Neural networks & Activation functions &  Loss functions \\
        \midrule
        Generator &    RELU and Linear & Binary Cross-entropy\\
        Discriminator &   RELU and Sigmoid & Binary Cross-entropy \\
        \bottomrule
	\end{tabular}
\end{table}

Out of the 500 generated samples, only 424 samples were observed to fall within the range of the training domain. Figure \ref{fig:GAN-samples-dist} presents the distributions of the generated PMP samples across the entire training domain. Even though correlations between certain PMPs can be observed in Figure \ref{fig:GAN-samples-dist}, it is important to note that the observed correlations among the generated samples are not constant. That is, the correlations vary each time new samples are generated. Hence, it does not imply that the learned distributions are correlated.

Figure \ref{fig:GAN-predictions} shows that the generated void fraction values closely resemble the magnitudes observed in the TRACE validation data. To evaluate how similar they are, the void fraction errors for all the samples are computed and the distributions of these errors are presented in Figure \ref{fig:GAN-error-distribution}. The error distributions of the four responses generally center around zero. The \texttt{VoidF1} error distribution has a smaller variance when compared to the other three axial locations, indicating a better accuracy. This is because \texttt{VoidF1} is calculated at the bottom of the rod. Figure \ref{fig:GAN-predictions} illustrates that most of the void fraction values at this specific location are close to zero. As a result, GANs find it relatively easier to learn the underlying data distribution. 

These results indicate that GANs have the capability to learn the underlying distributions within the training data and generate synthetic data that closely resembles the training data, without the need for a surrogate model to directly learn the input-output relationships. However, there is room for further improvement. This can be achieved by enhancing the network architecture, experimenting with different combinations of activation and loss functions, and perform automatic hyperparameters tuning with a hyperparameter optimization framework such as Optuna. One limitation of GANs is that the synthetic data is generated from random vectors, following the convention commonly used for image data generation in computer vision. Consequently, controlling the specific location or characteristics of the generated synthetic data becomes challenging. To address this issue, conditional GANs (CGANs) \cite{mirza2014conditional} can be employed, which alleviate this limitation by conditioning both the generator and discriminator on additional data. Additionally, another approach that can be explored to enhance GANs performance is Wasserstein GANs (WGANs) \cite{arjovsky2017wasserstein}, which offer improved stability and convergence properties compared to traditional GANs.

%%%%%%%%%%%%%%%%%%%%%%%%%%%%%%%%%%%%%%%%%%
\subsection{NFs results}
\label{sec:Results-NF}

The real NVP NFs model consists of 5 layers and was trained for 5000 epochs. Out of the 500 generated points, only 422 fell within the specified range for the PMPs. Figure \ref{fig:NF-samples-dist} illustrates the distribution of the generated samples across the domain, showing almost no correlations among the samples. After removing the points falling outside the domain, a validation process was conducted. The comparison between the void fraction values from TRACE simulations and the synthetic samples generated by the NFs model is shown in Figure \ref{fig:NF-predictions}. While the generated data agrees to a certain extent with the TRACE results, the deviations are overall larger than the GANs results. This is clearly shown in Figure \ref{fig:NF-error-distribution}. Even though the error distributions for the four responses are centered around zero, they have very large variances.

%Figure \ref{fig:NF-error-distribution} illustrates the error distributions that compare the generated data with the TRACE data. It is evident that the values generated for \texttt{VoidF2} exhibit the largest error values, with a few samples deviating significantly from the TRACE data. Although the PDF for \texttt{VoidF2} has a mean close to zero, it has a larger variance than the other three PDFs. On the other hand, the error behavior for \texttt{VoidF1} remains consistent with the GANs model, with relatively lower errors. However, in this case, the error values are relatively higher compared to the GAN models, ranging in $[-0.15,0.17]$. All four PDFs are centered around zero, but they display larger variances, indicating a moderate resemblance to the training dataset.
%When comparing the performance of the NFs model to that of the GANs model, it becomes apparent that both generative models are capable of producing synthetic datasets that exhibit a reasonable resemblance to the training data. However, the GANs model generally yields smaller errors compared to the NFs model. 

To enhance the performance of the NFs model and reduce errors in the generated data, improvements may be made by selecting better network structure. Additionally, another approach to enhance the NF model involves utilizing the non-linear independent component estimation method, which also incorporates a multi-scale architecture \cite{dinh2014nice}.

%%%%%%%%%%%%%%%%%%%%%%%%%%%%%%%%%%%%%%%%
\subsection{VAE results}
\label{sec:Results-VAE}

In the VAEs model, both the encoder and decoder networks were constructed with three fully connected layers, incorporating batch normalization and dropout techniques. The model was trained for 3000 epochs with a batch size of 30. Upon completing the training phase, the decoder network was utilized to generate 500 samples. Notably, all 500 generated samples were found to fall within the training domain. These samples are visually illustrated in Figure \ref{fig:VAE-samples-dist}. Figure \ref{fig:VAE-predictions} shows a very good agreement between the void fraction values generated by the VAEs model and those obtained from TRACE simulations. This is also demonstrated in Figure \ref{fig:VAE-error-distribution}, which shows that the errors for all four axial locations tend to cluster around zero, with very small variances.

%indicating a strong alignment between the generated values and the TRACE data. Additionally, the error range is relatively small, reflecting the accuracy of the generated samples.
%Among the four axial locations, \texttt{VoidF2} exhibits the largest error values, while \texttt{VoidF1} has the smallest errors. This discrepancy arises because \texttt{VoidF1} is relatively easier to learn due to a higher frequency of zero void fraction values at the bottom of the rod. Nonetheless, the variation in error magnitude between the four axial locations remains minor. All four error distributions exhibit a concentration around zero with narrow widths.
%These observations demonstrate the VAEs model's capability to generate samples that closely match the validation data, indicating agreement between the generated void fraction values and the results obtained from TRACE simulations.

It's worth mentioning that one of the limitations of the VAEs model is its inability to generate specific data points. This limitation can be addressed by utilizing the CVAEs model \cite{sohn2015learning}. CVAEs introduce an additional input to the model, representing the conditioning variable, which allows for more control over the generation process.

%%%%%%%%%%%%%%%%%%%%%%%%%%%%%%%%%%%%%%%%
\subsection{CVAEs results}
\label{sec:Results-CVAE}

The CVAEs model employs an identical network architecture to that of the VAEs, with the same number of epochs, learning rate, number of hidden layers, and number of neurons. The primary distinction lies in the input utilized. Rather than employing the nine-dimensional dataset for training the model, the PMP $\texttt{P1008}$ serves as the training labels (note that any other parameters will work as well). Consequently, when generating samples, random data is supplied alongside specific values of $\texttt{P1008}$ to generate data corresponding to the assigned values. The purpose here is to demonstrate the usage of CVAEs in generating specific data for scientific applications. That's why the first PMP $\texttt{P1008}$ was employed for demonstration purposes. After training the model, 500 points were generated by randomly selecting values for $\texttt{P1008}$. The data generation process yielded 492 points that fall within the designated range for the PMPs $[0.0,5.0]$.

The scatter plots in Figure \ref{fig:CVAE-samples-dist} display the generated data points for the PMPs. This figure illustrates the distribution of parameter values within the domain of the PMPs, highlighting their random generation. It demonstrates the desired characteristic of minimal correlation among the generated values for the five parameters. The 492 PMP values that fell within the specified domain were utilized to generate validation data through TRACE simulations. Figure \ref{fig:CVAE-predictions} presents a comparison between the void fraction values generated by the CVAEs model and those obtained from the TRACE simulations. The figure demonstrates a strong agreement between the generated values and the results from the TRACE simulations. This is verified in Figure \ref{fig:CVAE-error-distribution}, which shows that the errors in void fraction at the four axial locations are relatively small and concentrated. They exhibit a similar trend to the VAEs results, but with slightly narrower distributions.

By utilizing CVAEs, we were able to generate samples by specifying values for $\texttt{P1008}$. Such capability will be highly beneficial in situations where certain parameters need to be fixed while generating diverse samples by varying other parameters. In the present case, a single parameter was employed as a label, allowing us to select specific values for generating data based on that parameter ($\texttt{P1008}$). However, if the objective is to generate data by providing values for multiple parameters, it is essential to incorporate these parameters during the training of the model.

%%%%%%%%%%%%%%%%%%%%%%%%%%%%%%%%%%%%%%%%
\subsection{Comparisons and discussions}
\label{sec:Results-comparison}

In this subsection, our objective is to compare the four DGMs based on the quality of the generated samples. Table \ref{table:comparison-samples-generated} presents the number of samples that were generated within the designated domain for the PMPs. The results indicate that VAEs and CVAEs have the highest number of points falling within the desired domain. On the other hand, GANs and NFs exhibit a similar number of values falling within the domain. Similar results have been observed by repeating the data generation process a certain number of time. %This suggests that VAEs and CVAEs have effectively captured the distribution of the training data, although it does not necessarily imply that NFs and GANs have failed to do so. 

\begin{table}[ht]
	\footnotesize
    \captionsetup{justification=centering}
    \caption{Number of samples generated by the four DGMs that are inside the training domain.}
    \label{table:comparison-samples-generated}
    \centering
    \begin{tabular}{ c c c c c}
        \toprule
        DGMs & GANs & NFs & VAEs & CVAEs \\
        \midrule
        Generated samples inside the training domain & 424 & 422 & 500 & 492 \\
        \bottomrule
    \end{tabular}
\end{table}

Figure \ref{fig:predictions-comparison} shows that the VAEs and CVAEs models result in a higher level of agreement between the generated void fraction values and the TRACE simulated ones. The NFs samples exhibit more noticeable deviations. This behavior is more evident when examining the error distributions in Figure \ref{fig:error-distributions-comparison}. It is clearly noticeable that the NFs model exhibits a larger error range compared to the other three models. Moreover, the error distributions of the NFs model have a wider spread at all four locations. %Additionally, from this figure, we can observe that the error distribution for \texttt{VoidF2} is wider and has a larger error range for all four models. Conversely, \texttt{VoidF1} exhibits the smallest errors among the four locations for the four DGMs, as depicted by the narrower error distributions.

The statistics including the mean values ($\mu_{\text{error}}$) and standard deviations ($\sigma_{\text{error}}$) of the error distributions for the four models are summarized in Table \ref{table:error-comparison-DGMs}. While $\mu_{\text{error}}$ being close to zero indicate that, on average, the models are relatively unbiased, it is $\sigma_{\text{error}}$ that provide crucial information about the dispersion of errors for each model. Comparing the $\sigma_{\text{error}}$ allows for a more comprehensive assessment and comparison of the four models. 

\begin{table}[htbp]
	\footnotesize
	\captionsetup{justification=centering}
	\caption{Statistics of the void fractions errors by the four DGMs.}
	\label{table:error-comparison-DGMs}
	\centering 
	\begin{tabular}{l c c c c c c}
	   \toprule
          DGM & Statistics & \texttt{VoidF1} & \texttt{VoidF2} & \texttt{VoidF3} & \texttt{VoidF4}\\
          \midrule
        GANs & $\mu_{\text{error}}$ & -0.0002 & 0.0016 & 0.0015 & -0.0037\\
             & $\sigma_{\text{error}}$ & 0.0041 & 0.0117 &0.0085 & 0.0054\\
             \hline
        NFs  & $\mu_{\text{error}}$ & -0.0001 & -0.0072&0.0015 & -0.0001\\
             & $\sigma_{\text{error}}$ &0.0146 &0.0325 & 0.0274& 0.0202 \\
             \hline
        VAEs & $\mu_{\text{error}}$ & -0.0004 &-0.0071 & 0.0014& 0.0015\\
             & $\sigma_{\text{error}}$ & 0.0046  & 0.0128& 0.0076 & 0.0048\\
             \hline
       CVAEs & $\mu_{\text{error}}$ & -4.33e-05 &-0.0054 & 0.0018& 0.0015\\
             & $\sigma_{\text{error}}$ &0.0027  & 0.0108& 0.0057& 0.0040\\
        \bottomrule
	\end{tabular}
\end{table}

In accordance with the observations made from the fitted distributions of the errors, the $\sigma_{\text{error}}$ values for the NFs model were found to be the largest. This indicates that, in this particular application, the NFs model may be less suitable for scientific data generation. On the other hand, the GANs, VAEs, and CVAEs exhibit relatively close values with $\mu_{\text{error}}$ values close to zero and small $\sigma_{\text{error}}$ values. Among these models, the VAEs and CVAEs models have the best synthetic data quality.
%It is worth noting that \texttt{VoidF2} exhibits the same behavior across all the models, indicating that it was the most challenging location to learn. On the other hand, \texttt{VoidF1} is the easiest location to learn for all four models. This is likely due to \texttt{VoidF1} being measured at the lower part of the rod where values are consistently very close to zero. Consequently, the standard deviations for \texttt{VoidF2} are the highest among all the other locations, indicating greater dispersion in the error values. This suggests that predictions for \texttt{VoidF2} are more varied and less precise compared to the other locations. On the other hand, the values for \texttt{VoidF1} exhibit the smallest standard deviations, indicating that the errors are more consistent and less spread out around zero.

%%%%%%%%%%%%%%%%%%%%%%%%%%%%%%%%%%%%%%%%%%%%%%%%%%%%%%%%%%%%%%%%%%%%%%%%%%%%%%%%%%%%%%%%
%%%%%%%%%%%%%%%%%%%%%%%%%%%%%%%%%%%%%%%%%%%%%%%%%%%%%%%%%%%%%%%%%%%%%%%%%%%%%%%%%%%%%%%%
\section{Conclusions}
\label{sec:Conclusions}

DGMs are neural networks that can model and learn the underlying probabilistic distribution of a dataset. This enables them to generate synthetic data that aligns with the distribution of the training data. These models offer a potential solution to the common issue of data scarcity in nuclear engineering. With properly trained DGMs, it is possible to significantly expand the size of an existing dataset, thereby enabling DL algorithms to achieve better performance. In this paper, we tested the performance of several DGM techniques for data augmentation, namely GANs, real NVP NFs, VAEs and CVAEs. The DGMs were trained using TRACE simulation data derived from the BFBT benchmark. The dataset comprised 200 samples obtained from TRACE simulations, where the inputs consisted of five significant physical model parameters and the outputs were void fractions at four different axial locations. The deliberate intention in creating a small-sized training dataset was to mimic a scenario characterized by limited data availability. Subsequently, the trained DGMs were used to generate synthetic data aimed at resembling the original training dataset. To assess the plausibility of the synthetic data, we performed a validation step by comparing it with TRACE simulations conducted at the input values corresponding to the synthetic samples.

The synthetic data generated by the four models demonstrated their ability to produce credible samples. Among them, GANs, VAEs and CVAEs exhibited comparable and relatively smaller error values. NFs, on the other hand, yielded higher error values with larger variances for all the four void fractions. Additionally, VAEs and CVAEs had a greater number of values that fell within the training domain.
%Consistent with previous observations, the values generated for \texttt{VoidF2} displayed a wider error distribution compared to the other four locations, while \texttt{VoidF1} exhibited a narrower error distribution.  

In this demonstration, we assessed the quality of the generated samples by comparing them with TRACE simulations. However, in real-life applications where data originates from experiments, such a comparison may not be feasible. Previous attempts have been made to develop an evaluation metric for DGMs \cite{salimans2016improved}, but a consensus on the most effective metric in this context has not been reached. Establishing a robust evaluation metric for DGMs would enhance their reliability and trustworthiness in practical applications.

It is worth mentioning that the architectures of the aforementioned models were manually designed, which means that we might not be comparing them using their optimal architectures. A promising approach to address this issue is the utilization of automated neural architecture search (NAS) methods. Significant progress has been made in the image domain using NAS for both neural architecture search and GANs. An example of such a method is AutoGAN \cite{gong2019autogan} \cite{heusel2017gans}, which has demonstrated highly competitive performance when compared to GANs that were engineered by human experts.

In future work, we aim to apply these models to more complex datasets. Furthermore, we plan to enhance them by incorporating additional variants, such as conditional GANs and non-linear independent component estimation methods for NFs, as well as optimizing their network architectures. We also intend to explore other DGM techniques, including diffusion models, for the purpose of data augmentation. 
%In addition, we will also investigate suitable evaluation metrics for assessing the performance of DGMs. Furthermore, we will explore the application of automated neural architecture search methods to optimize the architectures of these models, including the aforementioned AutoGAN approach.

%%%%%%%%%%%%%%%%%%%%%%%%%%%%%%%%%%%%%%%%%%%%%%%%%%%%%%%%%%%%%%%%%%%%%%%%%%%%%%%%%%%%%%%%
%%%%%%%%%%%%%%%%%%%%%%%%%%%%%%%%%%%%%%%%%%%%%%%%%%%%%%%%%%%%%%%%%%%%%%%%%%%%%%%%%%%%%%%%
%%\section*{References}
\bibliography{./bibliography.bib}

\end{document}